%% file: main.tex
\pdfoutput=1

\documentclass[11pt]{article}

\usepackage[final]{acl}

\usepackage{times}
\usepackage{latexsym}
\usepackage{booktabs}

\usepackage[T1]{fontenc}

\usepackage[utf8]{inputenc}

\usepackage{microtype}

\usepackage{inconsolata}
\usepackage{graphicx}
\usepackage{multirow}
\usepackage{colortbl}
\usepackage{listings}
\usepackage{xcolor}
\usepackage{amsmath}
\usepackage{amssymb}
\usepackage{amsfonts}
\usepackage{url}
\usepackage{hyperref}
\usepackage{tabularx}
\usepackage{comment}

\definecolor{codegreen}{rgb}{0,0.6,0}
\definecolor{codegray}{rgb}{0.5,0.5,0.5}
\definecolor{codepurple}{rgb}{0.58,0,0.82}
\definecolor{backcolour}{rgb}{0.95,0.95,0.92}
\definecolor{purple}{RGB}{147, 112, 219} 
\definecolor{green}{RGB}{0, 128, 0}

\lstdefinestyle{mystyle}{
    backgroundcolor=\color{backcolour},   
    commentstyle=\color{codegreen},
    keywordstyle=\color{magenta},
    numberstyle=\tiny\color{codegray},
    stringstyle=\color{codepurple},
    basicstyle=\ttfamily\footnotesize,
    breakatwhitespace=false,         
    breaklines=true,                 
    captionpos=b,                    
    keepspaces=true,                 
    numbers=left,                    
    numbersep=5pt,                  
    showspaces=false,                
    showstringspaces=false,
    showtabs=false,                  
    tabsize=2
}

\definecolor{lightyellow}{rgb}{255, 255, 0}
\newcommand{\data}{\texttt{Suri}}

\title{\data: Multi-constraint Instruction Following for \\Long-form Text Generation}

\author{
    Chau Minh Pham \quad
    Simeng Sun\thanks{Now at NVIDIA} \quad
    Mohit Iyyer \\[0.5em] 
    University of Massachusetts Amherst\\
    \texttt{\{ctpham,simengsun,miyyer\}@cs.umass.edu}
}

\begin{document}
\maketitle
\input{sections/0-abstract}
\input{sections/1-intro}
\input{sections/2-data}
\input{sections/3-experiment}
\input{sections/4-automatic}
\input{sections/5-human}
\input{sections/6-related}

\input{sections/7-conclusion}
\input{sections/discussion}
\input{sections/8-acks}

\bibliography{custom}
\input{sections/appendix}

\end{document}

%% file: sections/0-abstract.tex
\begin{abstract}
Existing research on instruction following largely focuses on tasks with simple instructions and short responses. In this work, we explore multi-constraint instruction following for generating long-form text. We create \data, a dataset with 20K human-written long-form texts paired with LLM-generated backtranslated instructions that contain multiple complex constraints. Because of prohibitive challenges associated with collecting human preference judgments on long-form texts, preference-tuning algorithms such as DPO are infeasible in our setting; thus, we propose Instructional ORPO (I-ORPO), an alignment method based on the ORPO algorithm. Instead of receiving negative feedback from dispreferred \emph{responses}, I-ORPO obtains negative feedback from synthetically corrupted \textit{instructions} generated by an LLM. Using \data, we perform supervised and I-ORPO fine-tuning on Mistral-7b-Instruct-v0.2. The resulting models, \data-SFT and \data-I-ORPO, generate significantly longer texts ($\sim$5K tokens) than base models without significant quality deterioration. Our human evaluation shows that while both SFT and I-ORPO models satisfy most constraints, \data-I-ORPO generations are generally preferred for their coherent and informative incorporation of the constraints.\footnote{Code \& Data are available at \url{https://github.com/chtmp223/suri}}
    
\end{abstract}

%% file: sections/1-intro.tex
\section{Introduction}
\begin{figure*}[ht!]
    \centering
    \includegraphics[width=0.9\linewidth]{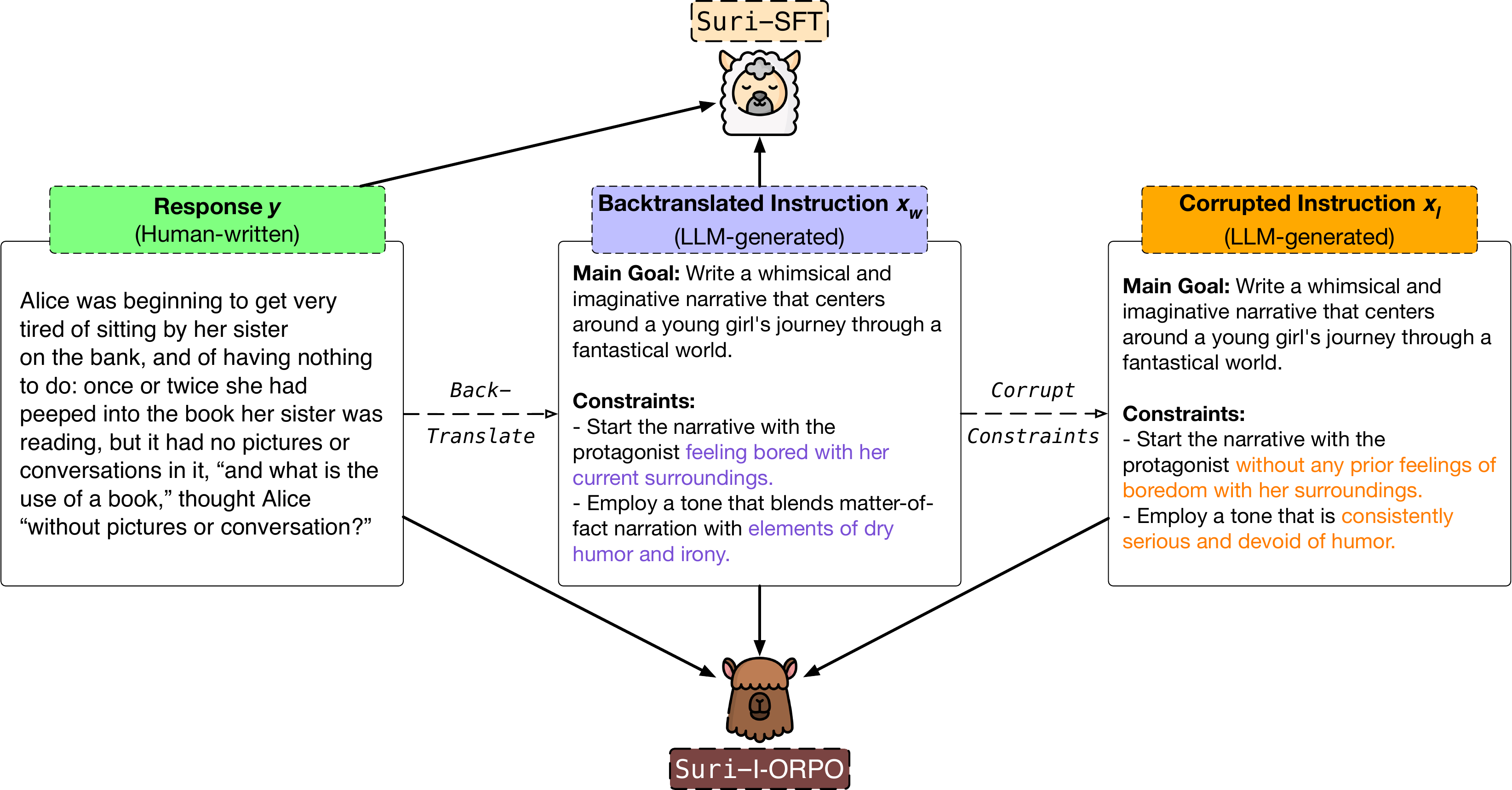}
    \caption{Our work consists of two stages. First, we construct the \data\ dataset using \textbf{\textcolor{green}{gold responses}} sampled from three existing datasets that include creative writing and open web text, along with \textbf{\textcolor{purple}{backtranslated instruction $x_w$}} and \textbf{\textcolor{orange}{corrupted instruction $x_l$}}. Second, we fine-tune Mistral-7B-Instruct-v0.2 on \data, resulting in two variations: \data-I-ORPO (via I-ORPO) and \data-SFT (via supervised fine-tuning).} 
    \label{fig:pipeline}
\end{figure*}
Improving the instruction-following abilities of modern large language models (LLMs) is critical to increasing their effectiveness and generalizability for many practical applications. However, most existing instruction-following datasets (e.g., Alpaca) contain only simple instructions that can be solved by short model generations~\citep{alpaca, DatabricksBlog2023DollyV2, köpf2023openassistant}. What about tasks with complex, multi-constraint instructions that can only be satisfied with \emph{long-form} outputs (i.e., thousands of tokens), such as creating detailed technical reports or writing engaging fictional narratives? 

We explore this question by conducting the first in-depth study of long-form instruction following with multi-constraint instructions. To facilitate our experiments, we create a new dataset, \data,\footnote{Suri is an alpaca breed known for its long, lustrous hair.} using instruction backtranslation~\citep{li2023self, koksal_longform_2023}. This process involves feeding a human-written long-form text (e.g., chapters from a novel) into an LLM to generate instructions that could have been followed to create the text. The resulting dataset, \data, consists of 20K texts paired with LLM-generated instructions, each containing $\approx$10 semantic and stylistic constraints (Figure \ref{fig:pipeline}).

How can we use \data\ to improve an LLM's long-form instruction following abilities? 
While supervised fine-tuning (SFT) has been quite effective for short-form datasets~\citep{mishra2022crosstask, wang2022supernaturalinstructions, sanh2022multitask, wei2022finetuned, chung2022scaling}, we observe that fine-tuned \data\ models often generate texts that are incoherent and fail to satisfy constraints towards the end in the instructions. Preference tuning methods such as DPO~\citep{rafailov2023direct} and RLHF~\citep{ouyang_training_2022} are challenging to use in this setting due to difficulties and cost in obtaining preference judgments on long-form texts~\citep{touvron2023llama,xu2023critical}. Specifically, when annotating preferences for long texts, human annotators may struggle to determine if different sections of the text are faithful to the instructions while simultaneously considering multiple aspects of the text, such as coherence and informativeness.

Motivated by this, we devise an alignment method that relies on synthetically \emph{corrupted} instructions. Specifically, we take the backtranslated instruction $x_w$ and corrupt its constraints using an LLM such that the gold response does not satisfy the corrupted constraints (for example, see $x_l$ in Figure \ref{fig:pipeline}). We then develop a variant of the Odds Ratio Preference Optimization objective~\citep[][ORPO]{hong2024orpo} to use these corrupted instructions as negative feedback. We refer to this alignment method as Instructional ORPO, or I-ORPO for short.

We conduct a series of automatic and human evaluations on generations from SFT and I-ORPO-tuned models to validate our method. Compared to the base model, Mistral-7b-Instruct-v0.2~\citep{jiang2023mistral}, both SFT and I-ORPO significantly increase the generation length from 1K to 5K tokens. Our fine-tuned models also improve the ability to differentiate between correct and corrupted instructions by at least 10\% while maintaining low levels of $n$-gram repetitions in the text. We find that LLM judges, such as GPT-4o~\citep{openai-gpt4o}, Gemini-1.5-Pro~\citep{geminiteam2024gemini15unlockingmultimodal}, and Claude-3.5-Sonnet~\citep{claude_sonnet}, cannot reliably evaluate long-form responses, which makes human evaluation crucial for assessing the constraint-following capabilities of our generations. Annotators note that our fine-tuned models effectively follow given constraints, with I-ORPO being preferred to our SFT model for its ability to incorporate constraints coherently, informatively, and enjoyably.

%% file: sections/2-data.tex
\section{The \data\ Dataset}
\begin{table*}[ht!]
\small
\begin{tabular}{p{2.5cm} p{4.75cm} p{0.5cm} p{4cm} p{0.75cm} p{1.0cm}}
\toprule
\textbf{Category} & \textbf{Dataset} & \textbf{Size} & \textbf{Domain} & \textbf{Prompt Length} & \textbf{Response Length} \\
\midrule
\multirow{2}{*}{\centering Writing Instructions} & ROCStory~\citep{mostafazadeh-etal-2016-corpus} & 50K & Creative Writing & 36 & 8 \\
&WritingPrompt~\citep{fan2018hierarchical} & 300K & Creative Writing & 28 & 735 \\
\midrule
Instruction-following & Alpaca~\citep{alpaca} & 52K & General Q\&As & 13 & 44 \\
\midrule
\multirow{3}{*}{\centering \parbox{3.5cm}{Long-form \\ Instruction-following}} & LongForm-C~\citep{koksal_longform_2023} & 28K & CommonCrawl, Wikipedia, StackExchange, Wikihow & 149 & 298 \\
&LongAlpaca-16K~\citep{chen2024longlora} & 12K & Science, Creative Writing, Wiki, General Q\&As & 5945 & 218 \\
& Scrolls~\citep{shaham2022scrolls} & 120K & Legal, Science, Entertainment & 33506 & 97 \\
\midrule
\parbox{3.5cm}{Multi-constraint \\Instructions} & Dolomites~\citep{malaviya2024dolomites} & 2K & 25 Academic Fields & 235 & 343 \\
\midrule
\rowcolor{lightyellow}\parbox{3.5cm}{Multi-constraint, \\Long-form \\Instruction-following} & Suri \textit{(this work)} & 20K & CommonCrawl, Creative Writing & 347 & 4371 \\
\bottomrule
\end{tabular}
\caption{\label{tab:data-comp} Comparison of \data\ with other single-turn datasets in terms of relevant data features, including writing instructions, instruction-following datasets, and constrained instructions. The data size as well as the average length of prompts and responses is either quoted from the original paper or calculated from publicly available subsets. \data\ is the only dataset featuring both constrained instructions and long responses (> 4K tokens) specifically designed for text generation.}
\end{table*}

We focus on the task of long-form writing, both fictional and non-fictional, under multiple constraints. When using an LLM for a complex writing task, users might have many constraints in mind and expect lengthy, detailed responses in the form of books, blog posts, etc. This task is particularly challenging for current LLMs, which struggle with generating coherent long-form outputs~\cite{guan2021long, wang2023language}, and this difficulty can be amplified when multiple constraints are involved. Recent instruction-following datasets have featured multi-constraint instructions~\citep{xu_kiwi_2024, malaviya2024dolomites} and long-form responses~\citep{koksal_longform_2023, chen2024longlora}, but none has integrated these two elements (Table \ref{tab:data-comp}). We bridge this gap by creating \data, which features complex instructions with multiple constraints and lengthy gold responses (2-5K words, about 3-6K tokens). 

We collect human-written English text samples, such as books, religious texts, and blog posts, to serve as gold responses ($y$). Since gathering human-written instructions for such lengthy responses is difficult and expensive, we turn to \emph{instruction backtranslation}~\citep{li2023self, koksal_longform_2023}, in which an LLM is provided with a human-written text (e.g., a short story) and prompted to generate instructions ($x_w$) that could have been followed to create that text. We further corrupt the constraints in $x_w$ to obtain synthetically corrupted instructions ($x_l$) for our I-ORPO alignment method. In total, \data\ contains 20K single-turn examples, each consisting of a backtranslated instruction $x_w$, corrupted instruction $x_l$, and a human-written response $y$. In this section, we detail our approach to selecting high-quality text samples ($\S$\ref{data:text}) and creating backtranslated instructions ($\S$\ref{data:instruction}). We also validate our generated instructions ($\S$\ref{data:human-eval}) and analyze the resulting dataset ($\S$\ref{data:analysis}).

\subsection{Collecting Responses}
\label{data:text}
Obtaining long-form gold responses $y$ through crowdsourcing or hiring experts requires significant cost and effort. As an alternative, we sample human-written texts in equal proportions from three existing datasets: ChapterBreak~\citep{sun2022chapterbreak}, Books3~\citep{presser, pile}, and RedPajama-Data-v2~\citep{together2023redpajama}. We truncate the sampled texts to between 2,048 and 5,024 words, making them significantly longer than those in existing instruction-following datasets (Table \ref{tab:data-comp}). The final \data\ dataset is divided into training, validation, and test sets in a 10K/5K/5K split.

\paragraph{ChapterBreak}
ChapterBreak (AO3 split) contains 7,355 fanfiction stories on Archive of Our Own (AO3), of which 6,656 texts are sampled for \data. We merge the individual chapters from the cleaned text into a single document.

\paragraph{Books3} Books3 contains 197K books on Bibliotik,\footnote{Due to copyright concerns, we only release the titles and IDs of the sampled data from this dataset. We provide a Python script to extract and clean the text so that users with access to Books3 can recreate the samples included in \data.} of which 6,698 texts are sampled. We use regular expressions to filter out irrelevant metadata, such as tables of contents and acknowledgments.

\paragraph{RedPajama-Data-v2} RedPajama contains over 100 billion documents from 84 CommonCrawl dumps. Unlike ChapterBreak and Books3, which consist primarily of books and literary narratives, RedPajama captures the style of everyday writing with informal textual content such as blog posts, obituaries, and more. We apply a set of quality filters (see Appendix \ref{appendix:quality-signals}) on the 2023-06 and 2023-14 snapshots to obtain a subset of $\approx300$K high-quality, non-duplicated documents written in English, from which 6,646 texts are sampled. 

\subsection{Creating Instructions via Backtranslation}
\label{data:instruction}
\data\ includes backtranslated instructions ($x_w$) and corrupted instructions ($x_l$). To create $x_l$, constraints from $x_w$ are minimally edited to be partially violated while still faithful to the overall main goal of the instruction. These corrupted instructions $x_l$, along with $x_w$ and $y$, serve as inputs for our I-ORPO preference tuning experiments.

\paragraph{Backtranslating Instructions} Our extracted gold responses do not come with accompanying instructions. Gathering these instructions can be costly and time-consuming, as annotators have to synthesize the instructions from long texts. Therefore, we use instruction backtranslation~\citep{li2023self, koksal_longform_2023} to generate the missing instructions. Specifically, we prompt GPT-4-turbo\footnote{GPT-4-turbo refers to \texttt{gpt-4-0215-preview}. Experiment done using temperature=0.6 and top\_p=0.9.} with a gold response $y$ to generate a corresponding instruction $x_w$ that contains a main goal, which summarizes the content of the text, and a list of $\approx$10 constraints (Table \ref{tab:backtranslate-prompt}). These constraints can focus on stylistic elements (how something is communicated through tone, language, sentence structure), semantic elements (what topics, meanings, and concepts are included), or a combination of both. Constraints can also be broad, applying to large portions of the text, or specific, addressing elements that occur only once.
The result is a set of highly detailed, multi-constraint instructions that cover different parts of the text ($x_w$ in Figure \ref{fig:pipeline}).

\paragraph{Corrupting Instructions} 
We want to use \data\ in our alignment experiments, which traditionally rely on preference judgments (e.g., labeled $y_w$ and $y_l$ pairs). However, obtaining these judgments for long-form outputs is challenging due to the many competing aspects to consider (e.g., faithfulness to instructions, overall coherence, etc.). Therefore, we focus on learning from a corrupted instruction $x_l$ instead of from $y_l$. To create $x_l$, we prompt GPT-4-turbo\footnote{Experiment done using model=\texttt{gpt-4-0125-preview}, temperature=0.0, top\_p=0.0 to ensure deterministic results.} to \emph{minimally} edit each constraint in $x_w$ while preserving the original main goal (Table \ref{tab:violate-prompt}). The resulting instructions average 386 tokens, closely matching the average length of gold instructions at approximately 411 tokens.\footnote{The first author manually reviews a random subset of 50 corrupted claims by comparing them to their corresponding original versions, and confirms that the corruptions are minimal and effective in invalidating the original constraints.}

\subsection{Validating Instructions}
\label{data:human-eval}
To validate whether the backtranslated instructions faithfully represent the original text, we conduct a human evaluation on a sample of 30 $(x_w, y)$ pairs. Three Upwork\footnote{See Appendix~\ref{appendix:human-eval} for recruitment and compensation details.} annotators are asked to read through the $(x_w, y)$ pairs, highlight all text spans in the response that support the given constraints, and determine if the response supports the instruction (Figure \ref{fig:labelstudio_cons}).

Our findings indicate that, on average, about 87\% of the constraints are fully satisfied, with the remaining constraints being partially satisfied (see Appendix \ref{appendix:human-eval} for agreement statistics). We conclude that the backtranslated instructions are generally faithful to the original text.

\subsection{Instruction Diversity}
\label{data:analysis}
\begin{figure}[h!]
    \centering
    \includegraphics[width=1\linewidth]{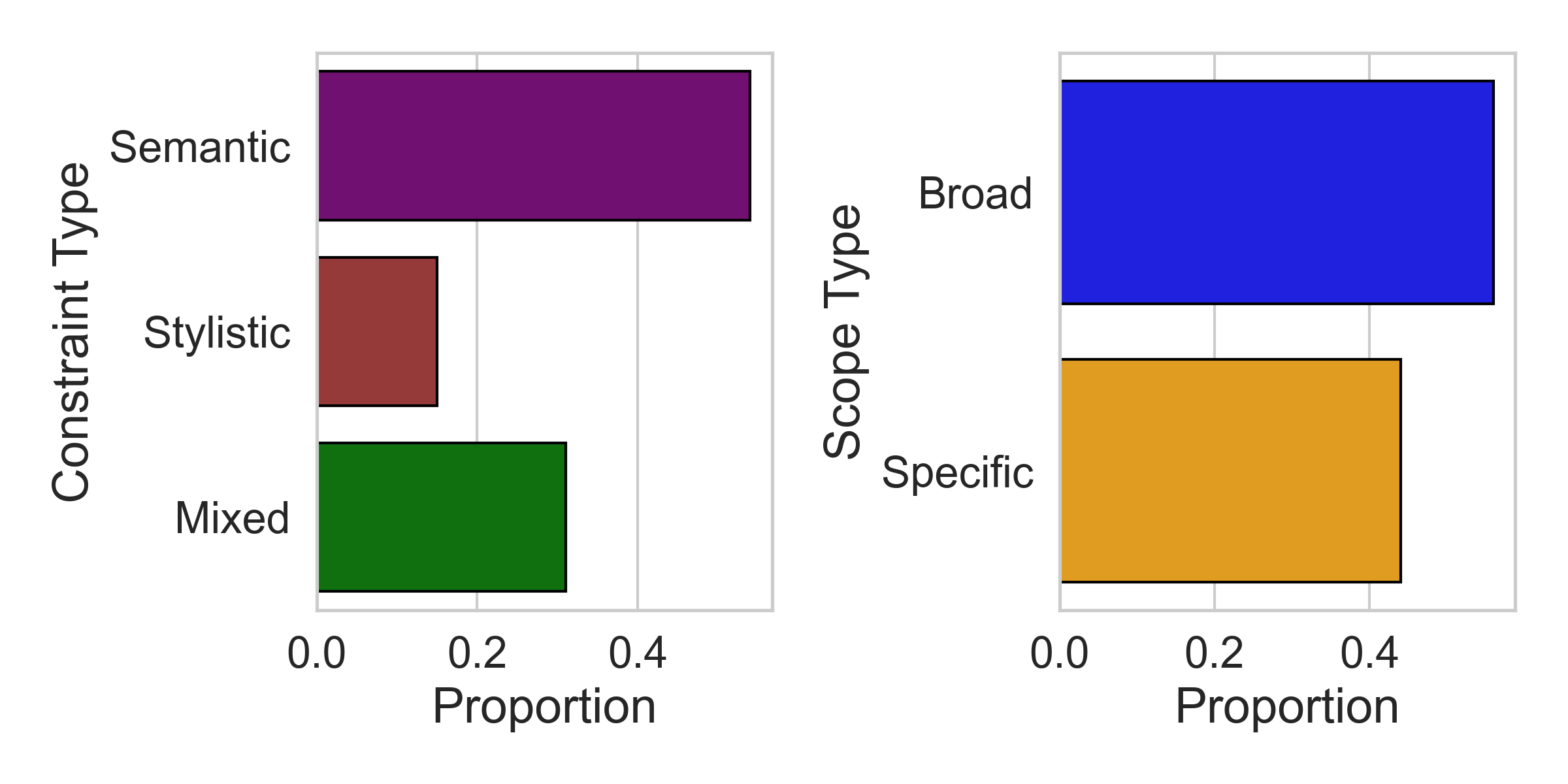}
    \caption{Average percentage of different constraint types within each instruction. The left figure categorizes the constraints based on their content, and the right figure refers to constraint scopes.}
    \label{fig:diversity}
\end{figure}

Instructions in \data\ focus primarily on long-form text generation, particularly crafting narratives or articles. Therefore, the key element that introduces diversity across these instructions is the accompanying list of constraints. Here, we measure the proportion of constraints being broad/specific or focusing on semantic/stylistic elements. We prompt Mistral-7B-Instruct-v0.2 \citep{jiang2023mistral},\footnote{Experiment done using greedy decoding. The first author manually verifies an output subset.} to assign each constraint to the applicable category. We find that semantic constraints account for more than half of each instruction, followed by mixed constraints (Figure \ref{fig:diversity}). Broad constraints, on the other hand, make up 56\% of the total constraints. Overall, the distribution of constraint types is relatively balanced, with a stronger emphasis on broad and semantic constraints.

%% file: sections/3-experiment.tex
\section{Aligning language models with \data}
Our goal is to assess whether \data\ helps improve the instruction-following capabilities of Mistral-7B-Instruct-v0.2 for long-form text generation. 

We experiment with two methods of fine-tuning Mistral-7B-Instruct-v0.2 on \data: supervised fine-tuning (SFT) using $(x_w, y)$ pairs and a modified ORPO alignment~\citep{hong2024orpo} using $(x_w, x_l, y)$ triplets. We emphasize that preference judgments are difficult to obtain for long-form responses due to numerous aspects of the text that must be considered with respect to the instructions. Therefore, we perform model alignment with correct instruction $x_w$ and corrupted instruction $x_l$ instead. Full details on fine-tuning libraries, hardware configurations, and hyperparameters can be found in Appendix \ref{appendix:modeling}.

\paragraph{\data-I-ORPO} 
Odds Ratio Preference Optimization (ORPO)~\cite{hong2024orpo} combines SFT and preference alignment by incorporating a log odds ratio term into the negative log-likelihood loss. We choose ORPO due to its simplicity, competitive performance with other preference tuning algorithms and the ease with which we can modify for our setting. 

The original algorithm learns from preference judgments, requiring access to chosen and rejected responses in the $(x, y_w, y_l)$ format. Since our dataset contains gold and corrupted instructions instead, we modify ORPO so that the algorithm accepts $(x_w, x_l, y)$. We refer to this modified method as Instructional Odds Ratio Preference Optimization (I-ORPO), where the modified loss is calculated as:
\begin{align}
    \mathcal{L}_{\text{I-ORPO}} = \mathbb{E}_{(x_w, x_l, y)} \left[ \mathcal{L}_{\text{SFT}} + \lambda \cdot \mathcal{L}_{\text{I-OR}} \right]
\end{align}
\text{where}
\begin{align}
    \mathcal{L}_{\text{I-OR}} = - \log \sigma \left(\log \frac{\textbf{odds}_{\theta}(y|x_w)}{\textbf{odds}_{\theta}(y|x_l)} \right) \\
    \textbf{odds}_{\theta}(y|x) = \frac{P_{\theta}(y|x)}{1- P_{\theta}(y|x)}
\end{align}

In the original ORPO formulation, the model is learning if the log probability of $P_\theta(y_w|x)$, denoted $\text{logps}(y_w|x)$, increases and log probability of $P_\theta(y_l|x)$, denoted $\text{logps}(y_l|x_w)$,  decreases after a number of training steps, resulting in the log odds ratio increasing. In I-ORPO, the same $y$ is used for both instruction types. Therefore, the model is learning if the log probabilities $\text{logps}(y|x_w)$ and $\text{logps}(y|x_l)$ diverge while $\text{logps}(x_w)$ and $\text{logps}(x_l)$ 

remain stable. We observe this trend in Figure \ref{fig:orpo-curve}. Loss derivation and analysis are in Appendix \ref{appendix:orpo}.  

\begin{figure}
    \centering
    \includegraphics[width=1\linewidth]{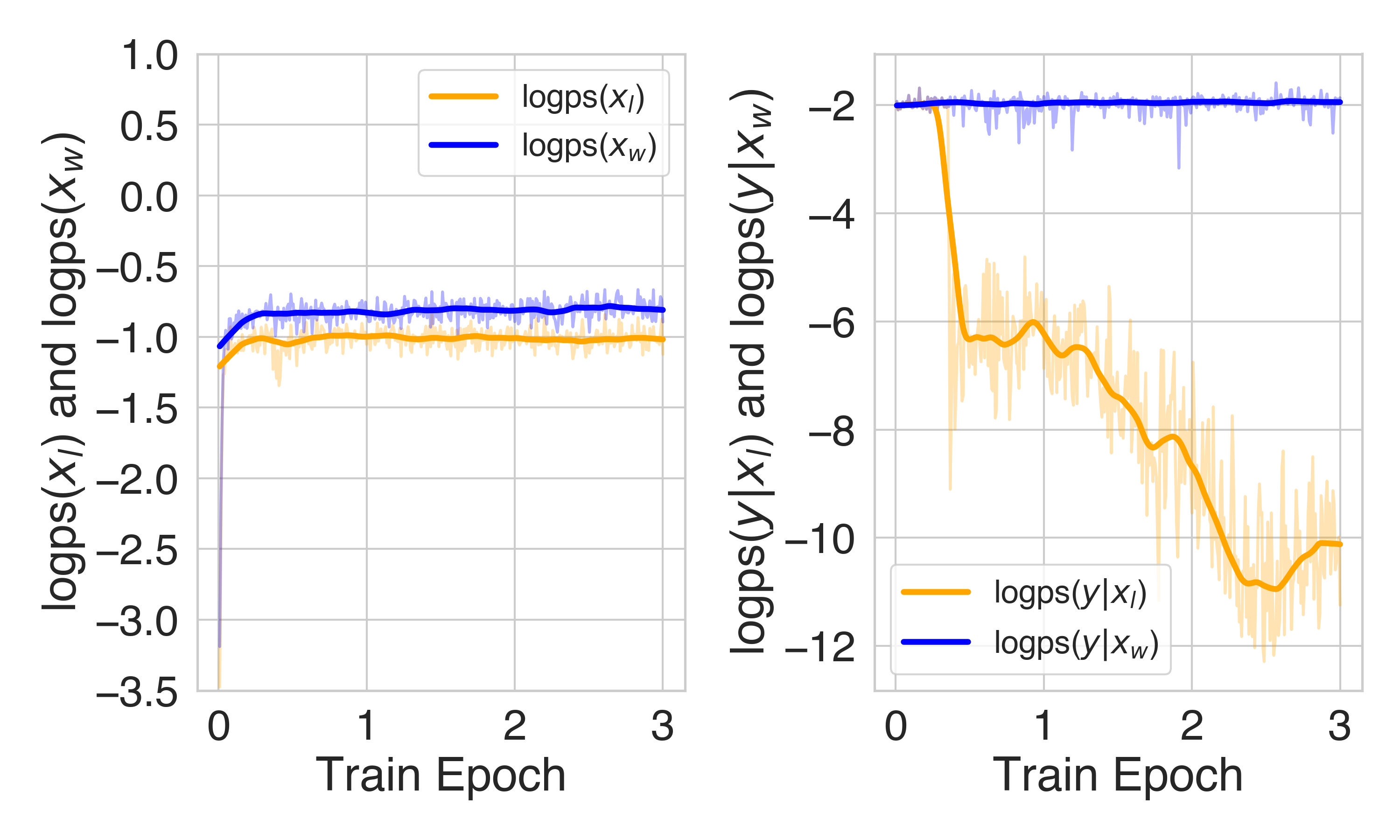}
    \caption{ORPO training curve. Left figure documents the log probability of the chosen and rejected prompts over 3 epochs. Right figure shows the log probability of the response given the chosen and rejected prompts over 3 epochs. A divergence between $\text{logps}(y|x_w)$ and $\text{logps}(y|x_l)$ is observed after 0.5 training epoch. 
    }
    \label{fig:orpo-curve}
\end{figure}

We perform I-ORPO fine-tuning with LoRA on Mistral-7B-Instruct-v0.2 for two epochs, using a learning rate of 5e-5, $\lambda$ of 0.4, and a LoRA rank and alpha of 16. We do not observe signs of the model learning with full-model tuning, so we choose to use LoRA fine-tuning instead. To minimize noise and improve the model's ability to distinguish between gold and corrupted instructions, we include a single constraint in each instruction, $x_w$ and $x_l$.

\paragraph{\data-SFT} We perform LoRA supervised fine-tuning~\cite{hu2021lora} on Mistral-7B-Instruct-v0.2 for two epochs using a learning rate of 5e-5, with a LoRA rank and alpha of 16. For each instruction $x_w$, we include a varying number of constraints to expose the model to different instruction formats. We do not use full-model tuning to match the I-ORPO training setting. 

For both \data-SFT and \data-I-ORPO, we set the number of epochs to 2, which is determined by a manual inspection of model generations at each epoch checkpoint. We observe that as the number of epochs increases from 1 to 5, both the response length and the number of 5-gram repetitions increase, indicating a trade-off between response length and repetition. After reviewing 30 generations at each epoch, we select the configuration that produces the most coherent responses and minimal repetition. Based on these criteria, the optimal number of epochs is 2, which balances the trade-off between response length and response quality.

%% file: sections/4-automatic.tex
\section{Automatic Evaluation}
Our automatic assessment demonstrates that both \data-I-ORPO and \data-SFT increase the length of the generated texts while maintaining a reasonable level of repetition. Compared to baseline models, \data-I-ORPO is more likely to assign higher log probabilities to tokens in the response given the correct instruction than the corrupted instruction.

\subsection{\data-I-ORPO and \data-SFT generate substantially longer text.}
We measure the average number of tokens\footnote{Measured using \texttt{tiktoken} package (\url{https://github.com/openai/tiktoken}) with ``o200k\_base'' encoding.}
in generations from our fine-tuned models (\data-I-ORPO and \data-SFT) and compare them to baseline models, including Mistral-7B-Instruct-v0.2, Llama-3-8B-Instruct~\citep{llama3modelcard}, and Mixtral-8x7B-Instruct-v0.1~\citep{jiang2024mixtral}. For faster inference, we use vLLM~\citep{kwon2023efficient} to generate outputs from the backtranslated instruction $x_w$.\footnote{Experiment done using greedy decoding, max\_token=10K. Inference prompts specify that 5K tokens should be generated.} Proprietary models like GPT-4 and Claude are excluded due to their maximum generation output limit of 4,096 tokens,\footnote{\href{https://docs.anthropic.com/en/docs/models-overview}{Claude documentation}; \href{https://platform.openai.com/docs/api-reference/audio}{OpenAI documentation}} whereas open-weight models allow for outputs of arbitrary maximum length.

\begin{figure}[ht!]
    \centering
    \includegraphics[width=0.9\linewidth]{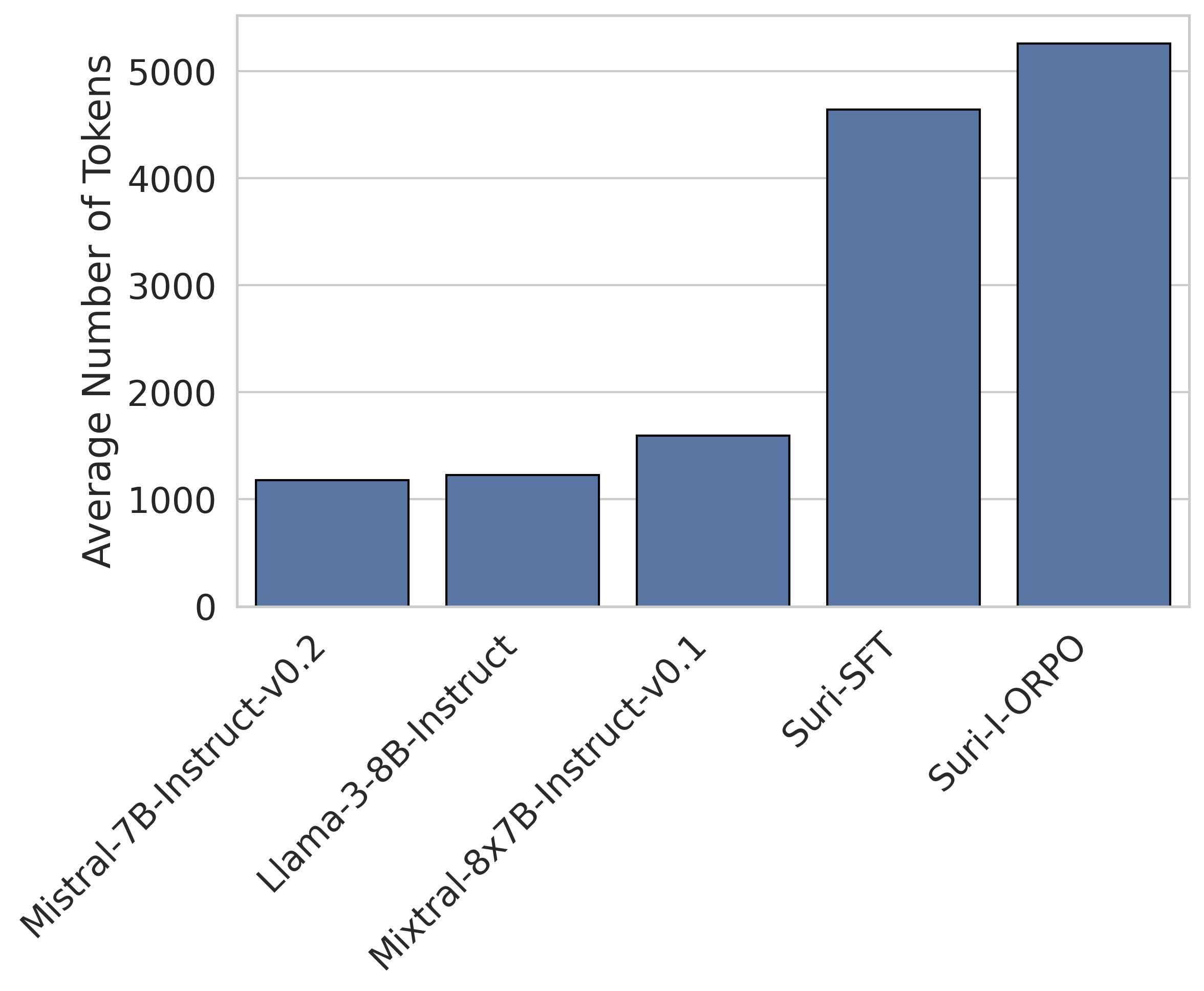}
    \caption{Average number of tokens in generations from baseline open-source models (Llama-3-8B-Instruct, Mixtral-8x7B-Instruct-v0.1, Mistral-7B-Instruct-v0.2) and our fine-tuned models (\data-I-ORPO, \data-SFT). }
    \label{fig:result-length}
\end{figure}

Our fine-tuned models, \data-SFT and \data-I-ORPO, generate significantly longer outputs compared to the open-weight baselines, with an average of approximately 4,800 and 5,100 tokens per generation, respectively (Figure \ref{fig:result-length}). These lengths exceed the maximum generation capacity of proprietary models, which is limited to around 4,096 tokens. Among the baselines, Mixtral produces the longest generations, averaging over 1,500 tokens, while Mistral-Instruct generates the shortest outputs, around 1,100 tokens per generation.

\subsection{\data-I-ORPO and \data-SFT do not degenerate into repetitions at longer sequences.}
We analyze the presence of repetitions in model generations. Since LLMs often degrade into repetitions over longer sequences, this measurement helps us identify when and how the model starts producing repetitive content. Previous work~\citep{li-etal-2016-diversity, see2019massively} measures unigram, bigram, and trigram repetitions. However, we are interested in sentence-level repetitions, such as when the same phrase is repeated in a dialogue at the start of each sentence. Therefore, we measure 5- and 10-gram repetitions to capture these higher-level patterns. We count a repetition when a specific $n$-gram appears at least three times in the text.

\begin{table}[h!]
\small
\centering
\begin{tabular}{p{1.25cm}p{0.75cm}p{0.75cm}p{0.75cm}p{0.75cm}p{0.75cm}}
\toprule
 & I-ORPO & SFT & Mistral-Instruct & Llama-Instruct & Mixtral-Instruct \\ \midrule
5-gram& 24\%& 29\%& 12\%& 26\%& 31\%\\
10-gram& 3\%& 3\%& 1\%& 2\%& 5\%\\ \bottomrule
\end{tabular}
\caption{Percentage of generations containing $n$-gram repetitions out of 5K generations from the test set (rounded to the nearest whole number).}
\label{tab:result-rep-perc}
\end{table}

\begin{figure}[ht!]
    \centering
    \includegraphics[width=0.8\linewidth]{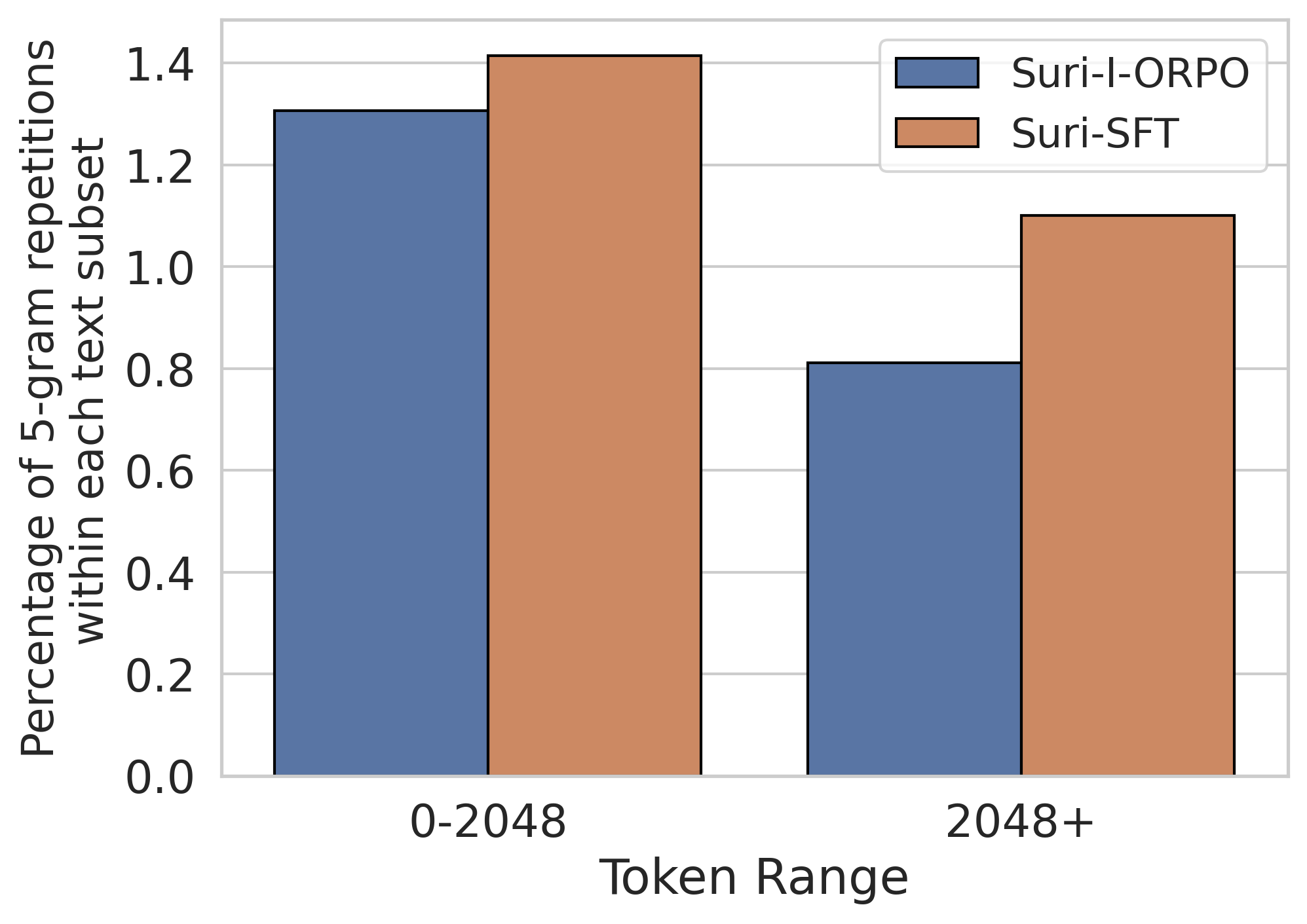}
    \caption{Average percentage of 5-gram repetitions before and after 2,048 tokens in each generation from I-ORPO and SFT models.}
    \label{fig:result-rep}
\end{figure}

Despite having the longest generations, \data-I-ORPO and \data-SFT maintain a low percentage of generations with $n$-gram repetitions (Table \ref{tab:result-rep-perc}). Among the baseline models, Mistral-Instruct has the lowest percentage of generations with repetition, possibly because its generations are also the shortest. Surprisingly, Llama-Instruct and Mixtral-Instruct, with their short generations, possess a greater proportion of generations with $n$-gram repetitions compared to our fine-tuned models. 

We further examine the percentage of 5-gram repetitions, normalized by the length of each text, generated by our fine-tuned models. As shown in Figure \ref{fig:result-rep}, the percentage of 5-gram repetitions does not increase after 2,048 tokens, indicating that our fine-tuned models do not exhibit degradation in longer sequences.

\subsection{I-ORPO improves ranking accuracy}
To understand the capabilities of models to differentiate between correct and corrupted instructions, we evaluate ranking accuracy~\citep{see2019massively, chen2024preference}. This involves measuring the percentage of cases in which the model assigns a higher probability to the gold response under the correct instruction than under the corrupted version. We calculate the sum of token log probabilities in the response given the previous tokens, denoted by $\text{logps}(y|x)$, and determine accuracy based on the proportion of times when $\text{logps}(y|x_w) > \text{logps}(y|x_l)$. A higher accuracy indicates that the model is more sensitive to the instructions and can determine which instruction is the correct instruction for the given response.

We use Hugging Face's Transformers~\cite{wolf-etal-2020-transformers} to access the probability distribution over vocabulary and measure the impact of instruction specificity on ranking accuracy across five different settings, which are defined by the number of all constraints included (M constraints in total) and the number of those included constraints that are corrupted: \texttt{(M,M)}, \texttt{(M,M/2)}, \texttt{(M,1)}, \texttt{(M/2,M/2)}, \texttt{(1,1)}. For example, in the \texttt{(M, M/2)} setting, both instructions include all constraints, but only half of the constraints are violated.

\begin{table}[ht!]
\small
\scalebox{0.99}{
\begin{tabular}{p{1.75cm}p{0.75cm}p{0.5cm}p{0.75cm}p{0.75cm}p{0.75cm}}
\toprule
Instruction Specificity & I-ORPO & SFT & Mistral-Instruct & Llama-Instruct & Mixtral-Instruct \\ \midrule
\texttt{(M,M)} & \textbf{100.0} & 99.8 & 90.6 & 65.7 & 66.5 \\
\texttt{(M,M/2)} & \textbf{100.0} & 99.2 & 92.1 & 57.5 & 60.4 \\
\texttt{(M,1)} & \textbf{98.3} & 91.0 & 90.4 & 47.7 & 55.2 \\
\texttt{(M/2,M/2)} & \textbf{99.9} & 97.8 & 79.7 & 60.0 & 57.4 \\
\texttt{(1,1)} & \textbf{98.4} & 81.2 & 62.5 & 50.9 & 48.5\\
\bottomrule
\end{tabular}}
\caption{Ranking accuracy on the \data\ test set across five levels of instruction specificity. Percentages are rounded to one decimal place.}
\label{tab:result-ranking}
\end{table}

\data-I-ORPO shows at least a 10\% improvement in ranking accuracy over the baseline Mistral-Instruct across all instruction specificity settings, with \data-SFT following closely (Table \ref{tab:result-ranking}). Mistral-Instruct remains a strong baseline, achieving the highest ranking accuracy among the three baseline models. In contrast, Llama-3-7b-Instruct and Mixtral-8x7b-Instruct perform the worst, trailing \data-I-ORPO by up to 50\%. We observe that settings with more constraints in the instruction, namely \texttt{(M,M)}, \texttt{(M,M/2)}, and \texttt{(M,1)}, generally lead to better performance. This trend suggests that seeing more constraints helps the model better differentiate between correct and corrupted constraints.

\subsection{LLM judges are unreliable for evaluating constraint satisfaction in long-form generation.}

We experiment with using LLMs to evaluate whether texts generated by our fine-tuned models follow the given constraints. Specifically, we provide GPT-4o~\citep{openai-gpt4o}, Gemini-1.5-Pro~\citep{geminiteam2024gemini15unlockingmultimodal}, and Claude-3.5-Sonnet~\citep{claude_sonnet} with a constraint and a generated text from our models and prompt it to determine whether the text fully satisfies, partially satisfies, or does not satisfy the constraint (Table \ref{prompt:llm-eval}).\footnote{Experiment done using OpenAI API for GPT-4o and Vertex API for Claude-3.5-Sonnet and Gemini-1.5-Pro. Temperature is set to 0.0 and the maximum number of generated tokens is set to 4096 for all models.} We then compare these results with judgments from three Upwork annotators on 30 texts generated by \data-SFT on the test set (obtained using the same procedure as in Section \ref{data:human-eval}). GPT-4o agrees with human annotators only 39\% of the time, with a significant 16\% disagreement between satisfaction and no satisfaction (Table \ref{tab:model-human-comparison}). Claude-3.5-Sonnet and Gemini-1.5-Pro lagging significantly behind, with Claude agreeing with humans only 24\% of the time, and Gemini 13\% of the time. Notably, Gemini refuses to annotate 39\% of the time, even when all safety filters have been disabled. We conclude that LLM judges do not align well with long-form human annotation, consistent with the findings of~\citet{xu_kiwi_2024} and~\citet{kim2024fables}.

\begin{table*}
    \small
    \centering
    \begin{tabular}{lccc}
        \toprule
        Category & GPT-4o & Claude-3.5-Sonnet & Gemini-1.5-Pro \\
        \midrule
        Agreement with human’s majority vote & 39\% & 24\% & 13\% \\
        Partial satisfaction - No satisfaction & 23\% & 6\% & 2\% \\
        Satisfaction - Partial satisfaction & 22\% & 63\% & 1\% \\
        Satisfaction - No satisfaction & 16\% & 7\% & 45\% \\
        \bottomrule
    \end{tabular}
    \caption{Types of agreement and disagreement between GPT-4o, Claude-3.5-Sonnet, Gemini-1.5-Pro, and human judges on 30 generations from \data-SFT.}
    \label{tab:model-human-comparison}
\end{table*}

%% file: sections/5-human.tex
\section{Human Evaluation}
\label{sec:exp-human-eval}
While our automatic assessments provide insights into the lexical information of the text, they do not capture its semantic content. Therefore, we conduct a human evaluation to determine if and how the constraints are satisfied by the outputs of \data-SFT and \data-I-ORPO. Human evaluation on 30 test set generations reveals that while both fine-tuned models satisfy constraints, \data-I-ORPO is preferred by humans for its ability to seamlessly incorporate the constraints into the final outputs.

\subsection{\data-I-ORPO and \data-SFT are effective at satisfying constraints.} Since GPT-4o judgments do not align with human annotations, we rely on human evaluation to determine how often \data-I-ORPO and \data-SFT follow the given constraints. This evaluation follows a similar setup as Section \ref{data:human-eval}, where annotators assess whether each constraint is satisfied, partially satisfied, or not satisfied by the generations. Two Upwork annotators complete 30 tasks, each containing a generation with around ten constraints, totaling 321 constraints. The generations are lengthy, averaging 4,000 words, and complex, with constraints spread throughout the text. Annotators spend approximately 20-25 minutes on each annotation and are paid \$200 in total for the task.

On average, \data-I-ORPO and \data-SFT meet most of the included constraints, achieving satisfaction rates of 67-68\% and partial satisfaction rates of 16-17\% (Table \ref{tab:human-eval-1}). Both models have the same proportion of unsatisfied constraints, accounting for 16\% of the total constraints. Annotators often note that narratives produced by \data-SFT contain inconsistent plot events and sometimes leave the narrative incomplete, resulting in some final constraints not being met. We attribute this behavior to the fact that some of the gold responses are truncated to between 2,048 and 5,024 words, which might omit the end of the original narrative. On the other hand, Mistral-I-ORPO produces narratives with coherent endings but can sometimes be too verbose, making it difficult to determine whether some constraints are satisfied.

\begin{table}
\small
\begin{tabular}{p{2.5cm}p{2cm}p{2cm}}
\toprule
 & \data-SFT & \data-I-ORPO \\ 
 \midrule
Satisfied & 67\% &  \textbf{68\%} \\
Partially Satisfied & \textbf{17\%}&  16\%\\
Not Satisfied & \textbf{16\%} & \textbf{16\%} \\ 
\bottomrule
\end{tabular}
\caption{Average percentage of satisfied constraints in \data-SFT and \data-I-ORPO generations. Percentages are rounded to the nearest whole number.}
\label{tab:human-eval-1}
\end{table}

\subsection{\data-I-ORPO are preferred over \data-SFT for coherent and informative constraint satisfaction.} In this evaluation, we are interested in how our fine-tuned models satisfy constraints in \data. We ask two annotators to compare text generations from \data-SFT and \data-I-ORPO with respect to a given constraint based on the following criteria:
\begin{itemize}
    \item \textit{Informativeness}: Which generation provides more details about the constraint? 
    \item \textit{Coherence}: Which generation effectively integrates the constraint with the rest of the text? 
    \item \textit{Readability/Enjoyability}: Which text sample is easier to read overall? 
\end{itemize}

The annotators also provide detailed justifications for their choices in each aspect of their judgments (see Figure \ref{fig:labelstudio_comp}).

\begin{table}[h!]
\centering
\small
\begin{tabular}{@{}ccc@{}}
\toprule
\textbf{Coherence} & \textbf{Informativeness} & \textbf{Enjoyability/Readability} \\ \midrule
72\% & 73\% & 67\% \\ \bottomrule
\end{tabular}
\caption{Win rate of \data-I-ORPO over \data-SFT in terms of coherence, enjoyability, and informativeness.}
\label{tab:sft-orpo}
\end{table}

Human annotators consistently prefer \data-I-ORPO to \data-SFT for about 60-70\% of the time across all three categories: coherence, informativeness, and enjoyability. Annotators note that \data-SFT often suffers from repetitive ideas, confusing plot points, and a lack of proper conclusions. In contrast, while \data-I-ORPO texts occasionally exhibit inconsistencies, they generally read more naturally, include interesting details, and are devoid of the robotic structure or flowery language often found in other LLM generations.

%% file: sections/6-related.tex
\section{Related Work}
\paragraph{Instruction Following Datasets} Open-ended instruction tuning involves fine-tuning LLMs to follow user instructions and generate high-quality text~\citep{wei2021finetuned, askell_general_2021, ouyang_training_2022, liu2023chain, rafailov2023direct}. Single-turn instruction-following datasets can be constructed by manual annotation, where instruction-response pairs are curated by humans~\citep{DatabricksBlog2023DollyV2,no_robots,zhou2024lima}. Another approach is distillation from proprietary LLMs, which can be done via techniques like Self-instruct~\citep{wang2023selfinstruct} to augment responses for each instruction~\citep{alpaca, xu2023wizardlm, xu2023baize}, Instruction Backtranslation to generate instructions given gold responses~\citep{koksal_longform_2023, li2023self}, or leveraging metadata to generate both instructions and responses~\citep{yin2023dynosaur}. While recent work has constructed instruction-following datasets with long-form responses~\citep{xiong_effective_2023, chen2024longlora, bai_longalign_2024} or multiple constraints~\citep{xu_kiwi_2024, zhou2023controlled, malaviya2024dolomites}, no prior effort has explored combining these two elements in single-turn instructions (see Table \ref{tab:data-comp}). \data\ is the first dataset to feature both complex instructions and long-form responses over 5k words. 

\paragraph{Alignment} Aligning language models with instruction-following data is crucial for ensuring that they respond to user instructions in a helpful and harmless manner~\citep{askell_general_2021, mishra2022crosstask, sanh2022multitask, chung2022scaling, wang2023far}. Popular preference tuning methods, such as RLHF, DPO, KTO, and ORPO~\cite{ouyang_training_2022,rafailov2023direct, ethayarajh2024kto, hong2024orpo}, achieve this by fine-tuning the models on human judgments of response quality~\citep{kreutzer-etal-2018-reliability, stiennon2022learning, ziegler2020finetuning, ramamurthy2023reinforcement}. However, collecting preferences for long-form responses is challenging due to the many competing aspects of the texts that need to be considered, such as instruction faithfulness and coherence~\citep{xu2023critical,kim2024fables, xu_kiwi_2024}, which prompts us to experiment with preference tuning on correct and correct instructions.

%% file: sections/7-conclusion.tex
\section{Conclusion}
In this work, we investigate the challenge of complex instruction following for generating long-form text. We introduce \data, a dataset of long human-written responses accompanied by backtranslated and corrupted instructions. We demonstrate the effectiveness of \data\ in improving the constraint-following capabilities of LLMs for long-form generation through supervised fine-tuning and I-ORPO. Human and automated evaluations show that our models generate high-quality, long-form responses while effectively satisfying constraints. 

%% file: sections/discussion.tex
\section*{Limitations}
\paragraph{Fine-tuning additional LLMs on \data} While we demonstrate the effectiveness of \data\ and I-ORPO on Mistral-7b-Instruct-v0.2, we have yet to experiment with fine-tuning other models on our dataset using I-ORPO.

\paragraph{Impact of surface features on I-ORPO} Even though I-ORPO works well on our dataset, we would like to explore how surface features, such as instruction length and the degree of information overlap between the instruction and response, affect its performance. 

\paragraph{Impact of truncating gold responses} In our experiments, we truncate gold responses to lengths between 2,048 and 5,024 words to make fine-tuning more cost-effective and computationally efficient. However, our released code includes an option that allows users to recover the full response text, and thus bypass the truncated version if needed.

\paragraph{Ranking accuracy on out-of-domain datasets} We report the ranking accuracy on the \data test set, where \data-SFT and \data-I-ORPO may have an advantage over the baseline models due to their fine-tuning on \data.     

\section*{Ethical Considerations}
Our human evaluation receives approval from an institutional review board. All annotators (US-based, fluent in English) gave their informed consent and participated with an hourly compensation of \$16, which meets the minimum wage in our state. Scientific artifacts are implemented according to their intended usage. 

%% file: sections/8-acks.tex
\section*{Acknowledgements}
We extend special gratitude to the Upwork annotators for their hard work and to the members of Unsloth, r/LocalLLaMA, and Together.ai community for helpful fine-tuning advice. We also thank Scott Niekum, Dzung Pham, and members of the UMass NLP lab for their insights on the project. This project was partially supported by awards IIS-2202506 and IIS-2312949 from the National Science Foundation (NSF) and an award from Open Philanthropy.

%% file: sections/appendix.tex
\appendix
\onecolumn
\section{Quality Filters for RedPajama-Data-v2}
\label{appendix:quality-signals}
Upon initial examination, we observe a significant presence of news and religious text in the corpus. Therefore, in addition to the following quality filters, we also downsample news and religious articles by excluding any article containing a source domain on our blocklist~\citep{ercexpo_us_news_domains_2023} or more than 0.05\% of words from a religious dictionary~\citep{ARDA_religiondictionary} to ensure the diversity of the gold responses. Table \ref{tab:my_label_part1} and \ref{tab:my_label_part2} show the quality filters used in RedPajama-Data-v2.  

\begin{table*}[htp!]
    \centering
    \begin{tabular}{@{}p{5.5cm}lp{6cm}p{1.5cm}} 
    \toprule
         Tags  &Values& Descriptions & Categories \\ 
         \midrule
         ccnet\_language\_score  & $> 0.65$ & score of the language identification model & CCNet \\
         \midrule
         ccnet\_perplexity  &$(35, 350)$ & perplexity of an LM trained on Wikipedia & CCNet \\
         \midrule
         rps\_doc\_books\_importance  &$ > 0$ & Given a bag of \{1,2\}-wordgram model trained on Books p, and a model trained on the source domain q, This is the logarithm of the ratio p(doc)/q(doc). & ML Heuristics \\
         \midrule
         rps\_doc\_curly\_bracket  &$0$ & The ratio between the number of occurrences of '\{' or '\}' and the number of characters in the raw text. Some pages inadvertently contained code. Since the curly bracket, "\{" appears in many programming languages (such as Javascript, widely used on the web) but not in natural text, we removed any pages that contained a curly bracket. & Natural Language \\
         \midrule
         rps\_doc\_frac\_no\_alph\_words  &$0.3$ & The fraction of words that contain no alphabetical character. & Natural Language \\
         \midrule
         rps\_doc\_lorem\_ipsum  &$0$ & The ratio between the number of occurrences of 'lorem ipsum' and the number of characters in the content after normalisation. & Natural Language \\
         \midrule
         rps\_doc\_unigram\_entropy  &$\geq3$ & The entropy of the unigram distribution of the content. This measures the diversity of the content and is computed using sum(-x / total * log(x / total)) where the sum is taken over counts of unique words in the normalised content. & Natural Language \\
            \bottomrule
         \end{tabular}
    \caption{Quality Signals used to filter RedPajamas Dataset - Part 1}
    \label{tab:my_label_part1}
\end{table*}

\begin{table*}[htp!]
    \centering
    \begin{tabular}{@{}p{5.5cm}lp{5.5cm}p{1.5cm}} 
    \toprule
    Tags & Values & Descriptions & Categories\\ 
    \midrule
    rps\_doc\_word\_count  &$(2048, 5024)$ & The number of words in the content after normalisation. & Natural Language \\
         \midrule
         rps\_lines\_javascript\_counts  &$0$ & The number of occurrences of the word "javascript" in each line. Many of the scraped pages contained warnings stating that Javascript should be enabled so we removed any line with the word Javascript. & Natural Language \\
         \midrule
    rps\_doc\_frac\_chars\_dupe\_10grams  &0.1 & \multirow{2}{6cm}{The fraction of characters in duplicate word ngrams.} & \multirow{2}{2cm}[0mm]{Repeti-tiveness} \\
         rps\_doc\_frac\_chars\_dupe\_5grams  &$0.15$ & &  \\
         rps\_doc\_frac\_chars\_dupe\_6grams  &$0.14$ & &  \\
         rps\_doc\_frac\_chars\_dupe\_7grams  &$0.13$ & &  \\
         rps\_doc\_frac\_chars\_dupe\_8grams  &$0.12$ & &  \\
         rps\_doc\_frac\_chars\_dupe\_9grams  &$0.11$ & &  \\
         rps\_doc\_frac\_chars\_top\_2gram  &$0.2$ & &  \\
         rps\_doc\_frac\_chars\_top\_3gram  &$0.18$ & &  \\
         rps\_doc\_frac\_chars\_top\_4gram  &$0.16$& \\
    \midrule
    rps\_doc\_ldnoobw\_words& $0$ & The number of sequences of words that are contained in the List-of-Dirty-Naughty-Obscene-and-Otherwise-Bad-Words blocklist. The blocklist is obtained from \url{https://github.com/LDNOOBW/}List-of-Dirty-Naughty-Obscene-and-Otherwise-Bad-Words & Sensitive / toxic content \\
    \midrule
    rps\_doc\_ut1\_blacklist & $0$ & A categorical id corresponding to the list of categories of the domain of the document. Categories are obtained from the UT1 blacklist. The list is obtained from \url{https://dsi.ut-capitole.fr/blacklists/}: {[}'adults', 'phishing', 'dating', 'gambling', 'filehosting', 'aggressif', 'ddos', 'mixed\_adult', 'chat', 'arjel'{]} & Sensitive / toxic content \\
    \bottomrule
    \end{tabular}
    \caption{Quality Signals used to filter RedPajamas Dataset - Part 2}
    \label{tab:my_label_part2}
\end{table*}

\newpage
\section{Prompts}
In this section, we show prompts to generate and analyze \data\ in Table \ref{tab:backtranslate-prompt}, \ref{tab:violate-prompt}, \ref{prompt:constraint-type}, \ref{prompt:constraint-scope}. Table \ref{prompt:llm-eval} shows the prompt used for our experiment with LLM judges. 

\label{appendix:prompt}
\begin{table*}[htp!]
\small
\begin{tabular}{p{0.95\linewidth}}
\toprule
\textbf{Prompt: Instruction Backtranslation/Reverse-engineering}
\\\midrule
Assume the author of the provided text followed a detailed set of instructions to produce their work. Your task is to infer what those original instructions may have been by composing your own set of instructions that could recreate key aspects of the given text. \\\\
Your response must include:
\begin{enumerate}
    \item An overarching instruction under the "Main Instruction" section that summarizes the goal of the instructions. 
    \item One bulleted list of specific constraints under the "Constraints" section that reflect the order of happenings/ideas in the original text. Constraints should focus on either stylistic elements (how something is communicated through tone, language, sentence structure), semantic elements (what topics, meanings, and concepts are included), or a combination of both. You should include specific elements from the text, but avoid using direct quotes. Aim for a fair balance of semantic, stylistic and mixed constraints. 
    \begin{itemize}
        \item Examples of stylistic constraints are "incorporate humor when discussing serious topics" or "use short, choppy sentences for emphasis."
        \item Examples of semantic constraints are "describe a supportive mother and absent father" or "mention an impressionist painting with a leopard."
        \item Mixed constraints blend stylistic and semantic elements, like "discuss impressionist art with an enthusiastic tone."
    \end{itemize}
\end{enumerate}

\#\#\# Document: \\
\{text\}\\\\

\#\#\# Your response: \\

\\\bottomrule
\end{tabular}
\caption{\label{tab:backtranslate-prompt} Prompt to reverse-engineer/backtranslate instructions. The placeholder \{text\} will be replaced with collected gold responses. Our instruction backtranslation experiment cost $\approx$ \$2K US dollars.}
\end{table*}

\begin{table*}[hp!]
\small
\begin{tabular}{p{0.95\linewidth}}
\toprule
\textbf{Prompt: Corrupt backtranslated instructions}
\\\midrule
You are given an instruction text that includes a main instruction and a list of constraints. Your task is to make minimal edits to violate each constraint. Your resulting constraints should be coherent with one another and also with the main instruction.\\\\

[Examples]\\
Main Instruction: Write a story on the life and death of Bob, who is a run-of-the-mill blue-collar worker in Texas, USA.\\
Constraints:
\begin{itemize}
    \item Use a first-person perspective that centers on the protagonist's perspective. $\rightarrow$ Use a third-person perspective that ensures a broad and neutral view of the narrative.
    \item Include cliffhangers at the end of each chapter to encourage readers to continue reading. $\rightarrow$ Do not include cliffhangers at the end of each chapter to encourage smooth readings.
\end{itemize}\\

[Provided Instruction]\\
\{instructions\}\\\\

When modifying the constraints, keep the following in mind:
\begin{enumerate}
    \item Ensure that your resulting constraints are coherent with one another and also with the main instruction. However, the original and modified constraints should be mutually exclusive and difficult to achieve simultaneously.
    \item Modify every constraint, but leave the main instruction unchanged.
    \item Your response should contain the original main instruction, followed by each original constraint and your minimally modified version. Format each constraint as: Original constraint $\rightarrow$ Your modified constraint.
\end{enumerate}

[Your response]

\\\bottomrule
\end{tabular}
\caption{\label{tab:violate-prompt} Prompt used to violate backtranslated instructions. The placeholder \{instructions\} are replaced with instructions that are produced with Prompt \ref{tab:backtranslate-prompt}.}
\end{table*}

\begin{table*}[hp!]
\small
\begin{tabular}{p{0.95\linewidth}}
\toprule
\textbf{Prompt: Assign constraint type (semantic, stylistic, mixed) to each constraint}
\\\midrule
You are a helpful assistant. You are given a constraint that you need to determine if it is a stylistic, semantic, or mixed constraint. Stylistic constraint emphasizes stylistic elements (how something is communicated through tone, language, sentence structure). Stylistic constraints focus on semantic elements (what topics, meanings, and concepts are included). Mixed constraints include both stylistic and semantic elements.\\\\

\#\#\# Examples: \\
Constraint: Incorporate humor when discussing the morbid, gut-wrenching scene of the protagonist's death. Use short, choppy sentences to create a sense of urgency and panic.\\
Your response: Stylistic\\\\

Constraint: The story must end with the protagonist's death in a car accident.\\
Your response: Semantic\\\\

Constraint: Using a first-person perspective, write a story on the life and death of Bob, a blue-collar worker in Texas, USA.\\
Your response: Mixed\\\\

Constraint: Include cliffhangers at the end of each chapter to encourage readers to continue reading.\\
Your response: Stylistic\\\\

\#\#\# Constraints: \\
Constraint: \{constraint\}\\\\

\#\#\# Your response: 

\\\bottomrule
\end{tabular}
\caption{\label{prompt:constraint-type} Prompt to assign constraint type (semantic, stylistic, mixed) to each constraint. The placeholder \{constraint\} will be replaced with a single constraint in each backtranslated instruction.}
\end{table*}

\begin{table*}[hp!]
\small
\begin{tabular}{p{0.95\linewidth}}
\toprule
\textbf{Prompt: Assign constraint scope (broad, specific) to each constraint}
\\\midrule
You are a helpful assistant. You are given a constraint that you need to determine if it is a specific or broad constraint. Specific constraints focus on an element that can be found in a specific part of the text. Broad constraints focus on an element that can be found throughout the text.\\\\

\#\#\# Examples: \\
Constraint: Throughout the narrative, use a first-person perspective that centers on the protagonist's perspective.\\
Your response: Broad\\\\

Constraint: Include cliffhangers at the end of the first chapter to encourage readers to continue reading. \\
Your response: Specific\\\\

Constraint: Introduce a new character in the middle of the story to add depth to the narrative.\\
Your response: Specific\\\\

Constraint: Include cliffhangers at the end of each chapter to encourage readers to continue reading.\\
Your response: Broad\\\\

\#\#\# Constraints: \\
Constraint: \{ $x_w$ constraint\}\\\\

\#\#\# Your response: 

\\\bottomrule
\end{tabular}
\caption{\label{prompt:constraint-scope} Prompt to assign constraint scope (broad/specific) to each constraint. The placeholder \{$x_w$ constraint\} is replaced with a single constraint from each backtranslated instruction.}
\end{table*}

\newpage
\section{Modeling Experiment Details}
\label{appendix:modeling}
All experiments are done using Flash-Attention 2~\citep{dao2023flashattention2}, DeepSpeed ZeRO 3~\citep{deepspeed}, PEFT~\citep{peft}, TRL library~\citep{vonwerra2022trl}, and Alignment Handbook~\citep{alignment_handbook2023}. Chat templates are as follows: 

\begin{lstlisting}[numbers=none]
    <|user|>
    {Instruction}</s>
    
    <|assistant|>
    {Response}</s>
\end{lstlisting}

The training configurations (Table \ref{tab:modeling-details}) are mostly similar for SFT and ORPO. We vary the learning rate (5e-4 to 5e-7), optimizer (8-bit vs. 32-bit), LoRA rank, and alpha (8 to 64), but none of these hyperparameters results in better generations.

\begin{table*}[h!]
\centering
\begin{tabular}{@{}ll@{}}
\toprule
\textbf{Configurations}         & \textbf{Values} \\ \midrule
Hardware (Training and Inference)  & 4xA100s        \\
Tracking                        & wandb          \\
\midrule
lora\_r                         & 16             \\
lora\_alpha                     & 16             \\
lora\_dropout                   & 0.05           \\
beta (for ORPO only)            & 0.4            \\
gradient\_accumulation\_steps   & 1              \\
gradient\_checkpointing         & True           \\
learning\_rate                  & 5.0e-5         \\
lr\_scheduler\_type             & cosine    \\
max\_length                     & 15024          \\
max\_completion\_length         & 15000          \\
max\_prompt\_length             & 5000           \\
num\_train\_epochs              & 2              \\
optim                           & adamw\_torch   \\
per\_device\_train\_batch\_size & 1 \\
\bottomrule
\end{tabular}
\caption{\label{tab:modeling-details}Training details for SFT and ORPO}
\end{table*}

\newpage
\section{I-ORPO Loss Derivation}
\label{appendix:orpo}
The derivation of $\mathcal{L_{\text{I-OR}}}$ closely resembles that of the original ORPO loss, with $d = (x_w, x_l, y) \sim D$. 
\begin{align}
\nabla_\theta \mathcal{L_{\text{I-OR}}} = \delta(d) \cdot h(d)
\end{align}
\begin{align}
\delta(d) &= \left( 1 + \frac{\textbf{odds}_\theta  (y|x_w)}{\textbf{odds}_\theta  (y|x_l)}\right)^{-1} \\
h(d) &= \frac{\nabla_\theta \log P_\theta(y|x_w)}{1-P_\theta(y|x_w)} - \frac{\nabla_\theta \log P_\theta(y|x_l)}{1-P_\theta(y|x_l)} 
\end{align}

The gradient of $\mathcal{L_{\text{I-OR}}}$ is the product of two terms: $\delta(d)$, which regulates the strength of parameter updates, and $h(d)$, which widens the contrast between $\text{logps}(y|x_w)$ and $\text{logps}(y|x_l)$. Specifically, as the odds ratio increases, $\delta(d)$ converges to 0. On the other hand, $h(d)$ has two gradients: $\nabla_\theta \log P_\theta(y|x_w)$, which minimizes $\log P_\theta(y|x_w)$, and $\nabla_\theta \log P_\theta(y|x_l)$, which maximizes $\log P_\theta(y|x_l)$. Additionally, $1-P_\theta(y|x_w)$ accelerates the update in the direction that maximizes $P_\theta(y|x_w)$. 
Following ORPO~\cite{hong2024orpo}, suppose that $g(x_w, x_l, y) = \frac{\textbf{odds}_\theta (y|x_w)}{\textbf{odds}_\theta (y|x_l)} $, we derive the loss as in \ref{eq:eq}.
\begin{figure*}
\begin{align}
\nabla_\theta \mathcal{L}_{I-OR} &= \nabla_\theta \log \sigma \left( \log \left( \frac{\textbf{odds}_\theta (y|x_w)}{\textbf{odds}_\theta y|x_l)} \right) \right)\\
&=\frac{1}{\sigma(\log g(x_w, x_l, y))}\cdot \nabla_\theta \sigma (\log g(x_w, x_l, y))\\
&=\frac{1}{\sigma(\log g(x_w, x_l, y))}\cdot \sigma (\log g(x_w, x_l, y))(1-\sigma (\log g(x_w, x_l, y))) \nabla_\theta \log g(x_w, x_l, y)\\
&=(1-\sigma (\log g(x_w, x_l, y))) \cdot \nabla_\theta \log g(x_w, x_l, y)\\
&=\sigma (-\log g(x_w, x_l, y)) \cdot \nabla_\theta \log g(x_w, x_l, y)\\           
&= \left( 1 + \frac{\textbf{odds}_\theta(y|x_w)}{\textbf{odds}_\theta(y|x_l)}\right)^{-1} \cdot \nabla_\theta \log\frac{\textbf{odds}_\theta(y|x_w)}{\textbf{odds}_\theta(y|x_l)}\\
&=\left( 1 + \frac{\textbf{odds}_\theta(y|x_w)}{\textbf{odds}_\theta(y|x_l)}\right)^{-1} \cdot \nabla_\theta \log\left(\frac{P(y|x_w)}{1-P(y|x_w)} \frac{1-P(y|x_l)}{P(y|x_l)}\right)
\end{align}
\\
$\nabla \log\left(\frac{P(y|x_w)}{1-P(y|x_w)} \frac{1-P(y|x_l)}{P(y|x_l)}\right)$ can be rewritten as:
\begin{align}
  &= \nabla_\theta \log\left(\frac{P(y|x_w)}{P(y|x_l)} \frac{1-P(y|x_l)}{1-P(y|x_w)}\right)\\
  &= \nabla_\theta \log\left(\frac{P(y|x_w)}{P(y|x_l)} \frac{1-P(y|x_l)}{1-P(y|x_w)}\right)\\
  &= \nabla_\theta \log \frac{P(y|x_w)}{P(y|x_l)} - (\nabla_\theta \log(1-P_\theta(y|x_w)) - \nabla_\theta \log(1-P_\theta(y|x_l)))\\
  &=\nabla_\theta \log \frac{P(y|x_w)}{P(y|x_l)} - \left(\frac{\nabla_\theta(1-P_\theta(y|x_w))}{1-P_\theta(y|x_w)} - \frac{\nabla_\theta(1-P_\theta(y|x_l))}{1-P_\theta(y|x_l)}\right)\\
  &=\nabla_\theta \log \frac{P(y|x_w)}{P(y|x_l)} - \left(\frac{-\nabla_\theta(P_\theta(y|x_w))}{1-P_\theta(y|x_w)} - \frac{-\nabla_\theta(P_\theta(y|x_l))}{1-P_\theta(y|x_l)}\right)\\
  &=\nabla_\theta \log \frac{P(y|x_w)}{P(y|x_l)} - \left(\frac{-P_\theta(y|x_w)\nabla_\theta \log P_\theta(y|x_w)}{1-P_\theta(y|x_w)} - \frac{-P_\theta(y|x_l)\nabla_\theta \log P_\theta(y|x_l)}{1-P_\theta(y|x_l)}\right)\\
  &= \nabla_\theta \log \frac{P(y|x_w)}{P(y|x_l)} - \left(-\textbf{odds}_\theta (y|x_w)\cdot\nabla_\theta \log P_\theta(y|x_w) + \textbf{odds}_\theta (y|x_l)\cdot\nabla_\theta \log P_\theta(y|x_l)\right)\\
  &= \nabla_\theta \log P_\theta(y|x_w) (1+\textbf{odds}_\theta(y|x_w)) - \nabla_\theta \log P_\theta(y|x_l) (1+\textbf{odds}_\theta(y|x_l))
\end{align}

The final equation is: 
\begin{align}
\label{eq:eq}
\begin{split}
    \nabla_\theta \mathcal{L}_{I-OR} &=\left( 1 + \frac{\textbf{odds}_\theta(y|x_w)}{\textbf{odds}_\theta(y|x_l)}\right)^{-1} \cdot (\nabla_\theta \log P_\theta(y|x_w)
    (1+\textbf{odds}_\theta(y|x_w)) -\\ &\nabla_\theta \log P_\theta(y|x_l) (1+\textbf{odds}_\theta(y|x_l)))
\end{split}\\
    &= \frac{1+\textbf{odds}_\theta(y|x_w)}{1+\frac{\textbf{odds}_\theta(y|x_w)}{\textbf{odds}_\theta(y|x_l)}} \cdot \nabla_\theta \log P_\theta(y|x_w) - \frac{1+\textbf{odds}_\theta(y|x_l)}{1+\frac{\textbf{odds}_\theta(y|x_w)}{\textbf{odds}_\theta(y|x_l)}} \cdot \nabla_\theta \log P_\theta(y|x_l)\\
    &=\left( 1 + \frac{\textbf{odds}_\theta  (y|x_w)}{\textbf{odds}_\theta  (y|x_l)}\right)^{-1} \cdot \left(\frac{\nabla_\theta \log P_\theta(y|x_w)}{1-P_\theta(y|x_w)} - \frac{\nabla_\theta \log P_\theta(y|x_l)}{1-P_\theta(y|x_l)}\right)
\end{align}
\end{figure*}

\newpage
\section{Preference Prompting}
In this evaluation, we provide the model with the gold response $y$ and both instructions $x_w$ and $x_l$. We then prompt the model to choose the instruction most relevant to the gold text, following~\citet{bai2022constitutional} and~\citet{lee2023rlaif}. The model should output `1' if the first instruction generates the text and `2' otherwise (Table \ref{tab:eval-prompt}). Next, we compare the log probabilities of the model outputting `1' and `2'. If the log probability for `1' is higher, we assume the model prefers whichever instruction came first in the prompt. The performance metric is determined by how often the model prefers the correct instruction, regardless of the order in which the correct instruction is presented. We experiment with Mistral-7b-Instruct-v0.2, \data-I-ORPO, \data-SFT, Mixtral-8x7b-Instruct-v0.1, Llama-3-7b-Instruct. All experiments use the Huggingface implementation with greedy decoding.
\begin{table*}[tp!]
\small
\begin{tabular}{p{0.95\linewidth}}
\toprule
\textbf{Prompt: \texttt{p(preference|prompt)} evaluation}
\\\midrule
You are an expert instruction rater. You will be given a text and two instructions, one of which is used to generate the text. Read through the text carefully, then determine which of the two instructions was used to generate the text. Answer only with "1" if the first instruction is correct, or "2" if the second instruction is correct. DO NOT give any reasoning. \\\\

\#\#\# Text:\\ 
\{text\}\\\\

\#\#\# First Instruction: \\
\{ins1\}\\\\

\#\#\# Second Instruction: \\
\{ins2\}\\\\

Which instruction is correct? Answer only with "1" if the first instruction is correct, or "2" if the second instruction is correct. DO NOT give any reasoning. \\\\

Your response: \\
\\\bottomrule
\end{tabular}
\caption{\label{tab:eval-prompt} Prompt used in the \texttt{p(preference|prompt)} evaluation. The \{text\} placeholder is replaced with gold responses, while the placeholders \{ins1\} and \{ins2\} are replaced with the correct and corrupted instructions, respectively. To mitigate any potential ordering bias, the order of the correct and corrupted instructions is shuffled. We will consider a response correct only if the model chooses the correct instruction, regardless of the ordering.}
\end{table*}

We observe that all models suffer from ``first instruction bias", where the model always outputs the first instruction as the correct instruction, regardless of whether that instruction is actually $x_w$ or not. 

\newpage
\section{Human Evaluation}
\label{appendix:human-eval}
\subsection{Recruitment}
We recruit human annotators, all of whom are fluent in English, from Upwork (\url{https://www.upwork.com}) for our human evaluation. Each task is assigned to two annotators, except for Instruction Validation, which involves three annotators. Annotators are compensated at a rate of \$16 per hour and generally work an average of 12 hours per task. All annotators have signed consent forms, and our study has been approved by our institutional review boards (IRB).

\subsection{Annotation}
\begin{figure*}[ht!]
    \centering    \includegraphics[width=0.95\linewidth]{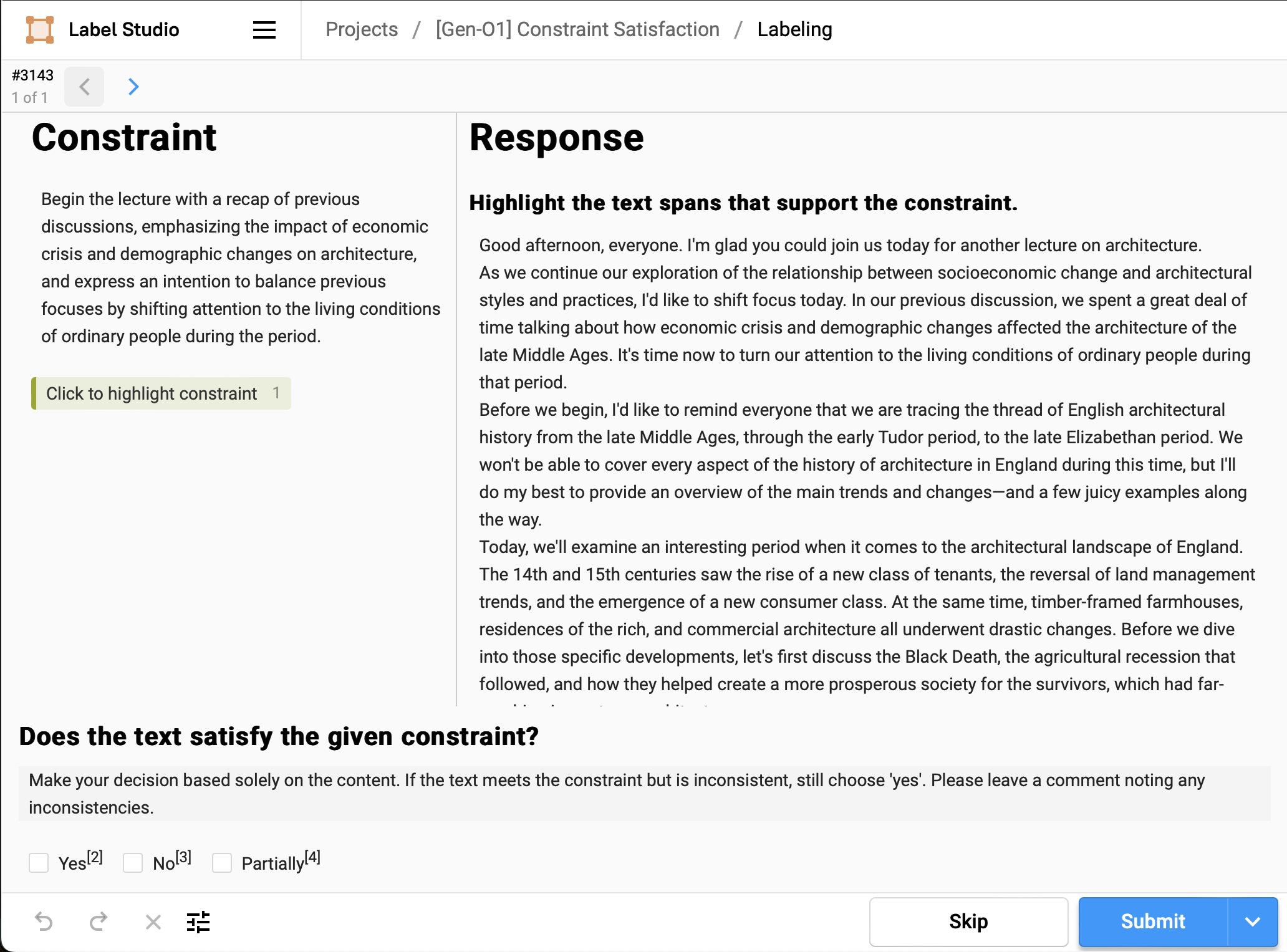}
    \caption{LabelStudio interface for annotating the validity of instructions. Annotators begin by carefully reading through the provided constraint and highlighting all the relevant text spans in the response supporting the constraint specified in the instruction. They then indicate whether the highlighted text satisfies the given constraint in the follow-up question.}
    \label{fig:labelstudio_cons}
\end{figure*}

\begin{figure*}[hp!]
    \centering    
    \includegraphics[width=0.95\linewidth]{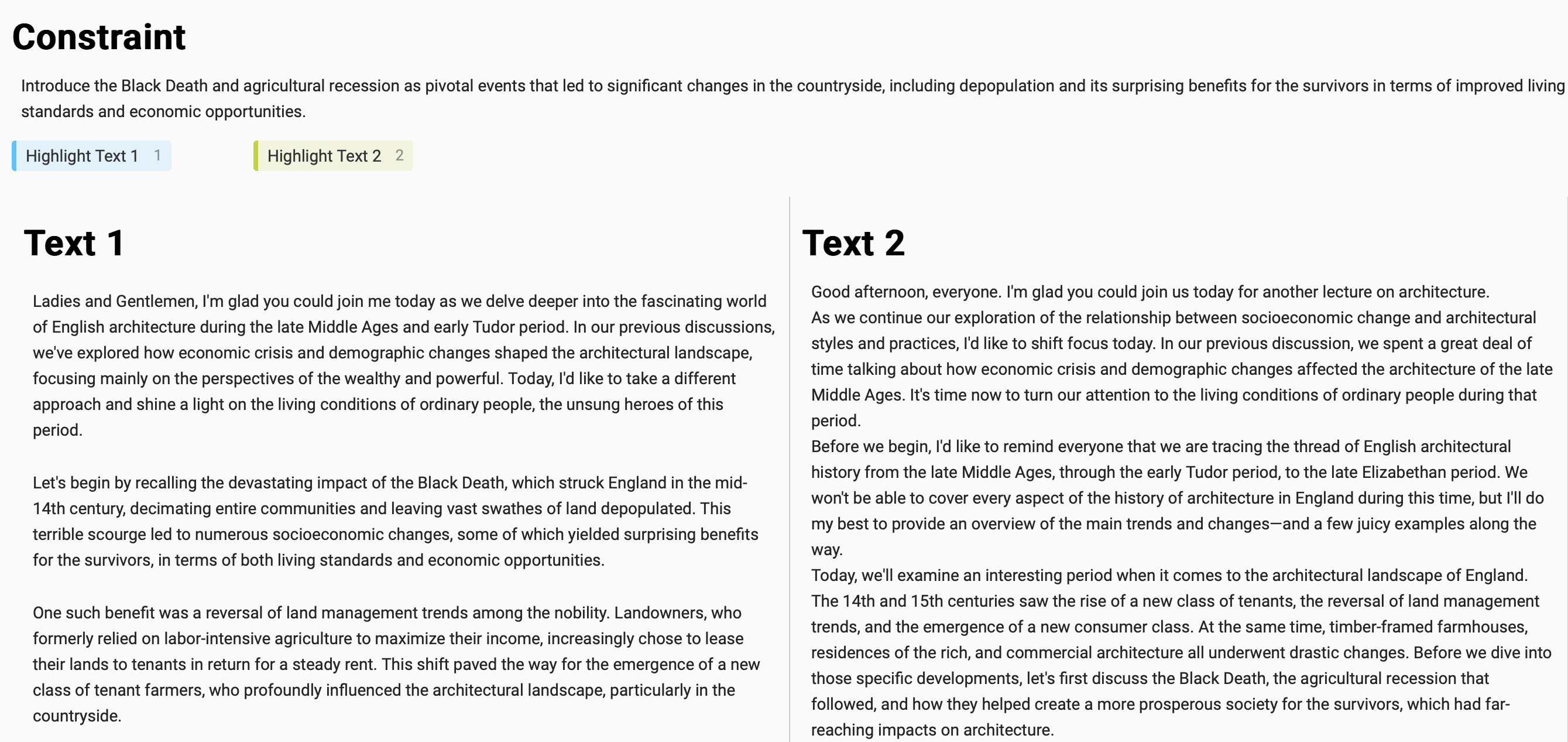}
    \\
    \includegraphics[width=0.95\linewidth]{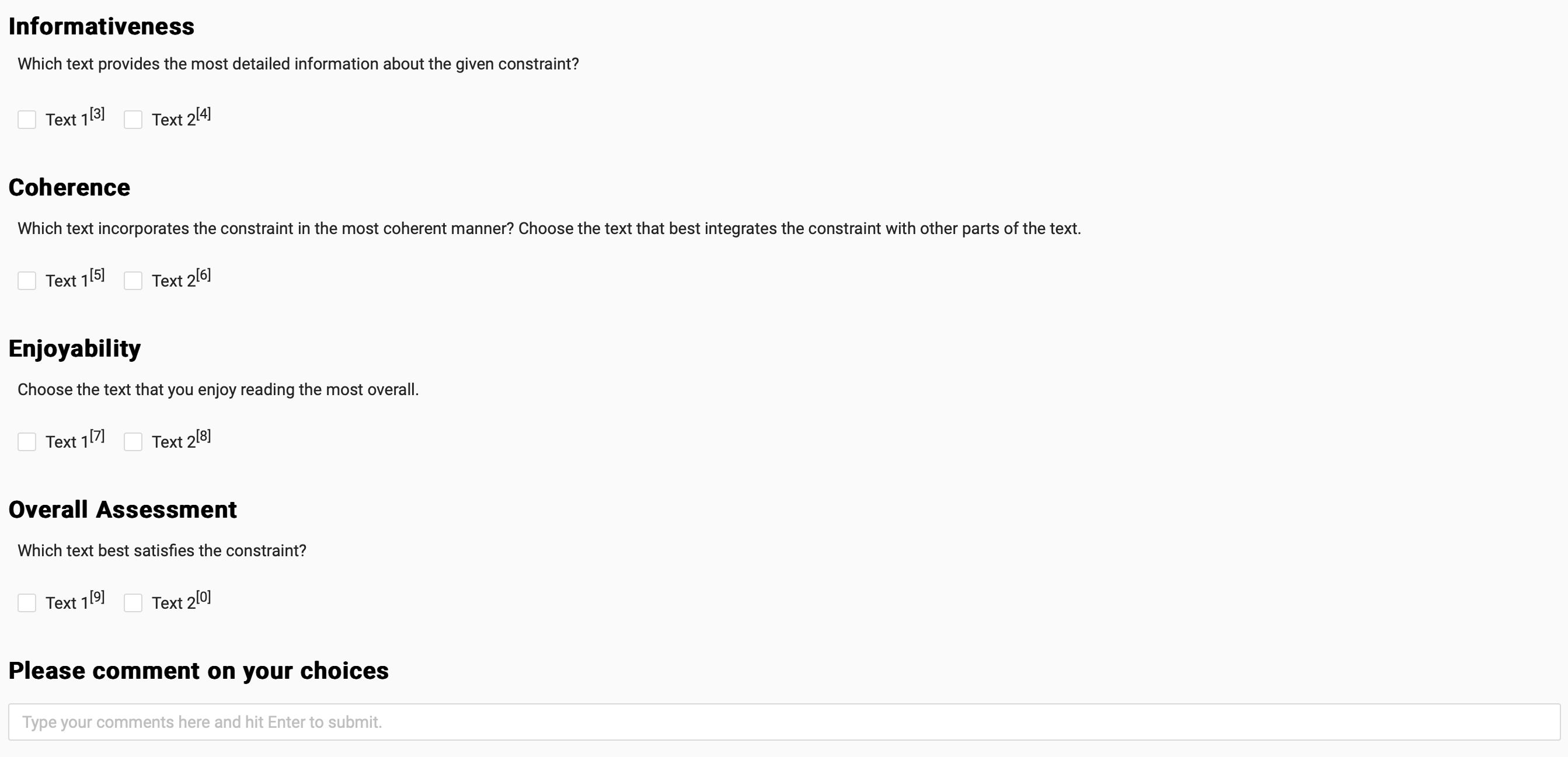}
    \caption{LabelStudio interface for comparing generated text. Annotators begin by carefully reading through the provided constraint. They then highlight all the relevant text spans in the response that support the constraint specified in the instruction. After that, annotators answer questions on the informativeness, enjoyability, and coherence of the provided texts. We shuffle the generations in each task to prevent bias.}
    \label{fig:labelstudio_comp}
    
\end{figure*}

Figure \ref{fig:labelstudio_cons} shows the LabelStudio interface for annotating instruction validity/constraint satisfaction. Figure \ref{fig:labelstudio_comp} features the interface for comparing text generations based on how they satisfy a given constraint. Annotators note that the interfaces are user-friendly.

\subsection{Annotator agreement in the instruction validity and constraint satisfaction evaluation}
\subsubsection{Power Analysis}
We conduct a power analysis~\citep{card-etal-2020-little} on our human evaluation data to estimate the number of text generations needed to achieve a power of 0.80 with a significance cutoff of 0.05 for our evaluation.
\begin{enumerate}
    \item Evaluation 1 - Annotators rate text on constraint satisfaction (yes, no, partially): For this task, we estimated the effect size using Cohen's w and employed a Chi-square goodness of fit test to determine the necessary number of text samples. We found that only 14.99 samples are needed to achieve a power of 0.80. In addition, with our current sample size of 30, we achieve a power of 0.9820, which exceeds the acceptable threshold of 0.80.
    \item Evaluation 2 - Annotators indicate preference between text from model 1 and model 2: For this task, we estimated the effect size using Cohen's h and used a z-test to calculate the required number of text samples. We determined that 23.17 samples are needed to reach a power of 0.80. With our sample size of 30, we achieve a power of 0.8902, which is above the acceptable threshold of 0.80.
\end{enumerate}

While we evaluate 30 text samples for the first task, each text sample is evaluated with respect to approximately 10 constraints from the instructions, as we aim to account for constraints towards the end that are often missed by the models. Therefore, each annotator must evaluate a total of 321 (constraint, text) samples. This process is costly, with each annotator compensated \$200 and taking $\approx 15$ hours to complete the task.

\subsubsection{Annotators' Agreement}
We note that Krippendorff's Alpha remains low across evaluation tasks, suggesting little to no agreement among the annotators. We attribute this pattern to the fact that our generations are long ($\approx$4k words on average), making it hard for annotators to follow the narrative sometimes. Final statistics reported in the paper is averaged between the annotators.

Table \ref{tab:appendix-agreement} further shows disagreement types for the instruction validity and constraint satisfaction evaluation. 
\begin{table*}[hp!]
\small
\centering
\begin{tabular}{p{6.5cm}p{2cm}p{1.75cm}p{2cm}p{1.5cm}}
\toprule
\textbf{Types} &\textbf{Krippendorff's alpha}& \textbf{Satisfied vs Partially Satisfied}& \textbf{Partially Satisfied vs Not Satisfied}&\textbf{Satisfied vs Not Satisfied}\\
\midrule
Instruction Validation (Section \ref{data:human-eval})  &0.1& 0.30 &  0.03&0.0\\
Constraint Satisfaction - \data-SFT (Section \ref{sec:exp-human-eval})  &0.0& 0.52&  0.24& 0.24\\
Constraint Satisfaction - \data-I-ORPO (Section \ref{sec:exp-human-eval})  &0.2& 0.60&  0.34& 0.06\\
I-ORPO vs SFT - Coherence (Section \ref{sec:exp-human-eval})  &0.0& -& -& -\\
I-ORPO vs SFT - Informativeness (Section \ref{sec:exp-human-eval})  &0.0& -&  -& -\\
I-ORPO vs SFT - Enjoyability (Section \ref{sec:exp-human-eval})  &0.1& -&  -& -\\
\bottomrule
\end{tabular}
\caption{\label{tab:appendix-agreement} Types of disagreement among annotators in the instruction validation and constraint satisfaction tasks. Most disagreements arise over whether the text fully or partially satisfies the constraints.}
\end{table*}

\begin{table*}[ht!]
\small
\begin{tabular}{p{0.95\linewidth}}
\toprule
\textbf{Prompt: LLM evaluation}\\
\midrule
You will be given a text and its corresponding instruction, which contains the text's main goal and a constraint. Determine whether the text satisfies the constraint (not the main goal). You should return your answer (Yes/No/Partially) along with your reasoning and a quote in the text that supports your reasoning (the quote should not contain any double quotation marks). Your answer should contain 3 fields: "answer", "reasoning", and "quote". DO NOT output anything else other than the response, which starts with "<<" and ending with ">>".\\\\

\# Example 1: The text satisfies the constraint. \\
- Main goal: Write a first-person narrative describing a serene morning in a remote village.\\
- Constraint: You must not use the letter 'e'. \\
- Text: Dawn cracks with a yawn. On a hill, a hut sits, tranquil. Bright light climbs, casting gold on grass. In this calm morning, air is cool, birds sing softly. I stroll down paths, sipping hot cocoa, watching day start. Such is this dawn's charm, lifting spirits, as world awakens.\\
- Your response: <<"answer": "Yes", "reasoning": "The text does not contain any 'e', which satisfies the constraint.", "quote": "Dawn cracks with a yawn...">>\\\\

\# Example 2: The text does not satisfy the constraint. \\
- Main goal: Compose a narrative that takes place entirely within the confines of a single, small room. \\
- Constraint: The story must not include any direct interaction or communication with other characters, whether through dialogue, notes, or any form of digital communication.\\
- Text: Sarah sat quietly in the corner of the small, dimly lit library room, surrounded by towering bookshelves filled with dusty volumes. Her focus was broken by a soft knock on the door. "Sarah, are you there?" her friend Emily's voice called out gently from the other side. Sarah, startled yet relieved to hear a familiar voice, responded, "Yes, I'm here, Emily. Just give me a moment, I'll open the door." They spent the next hour talking about the books Sarah had been reading and their plans for the weekend, making the small room feel a lot less lonely.\\
- Your response: <<"answer": "No", "reasoning": "The text includes a dialogue between Sarah and Emily, while the constraint specifies that the story must not include any direct interaction.", "quote": "'Sarah, are you there?' her friend Emily's voice called out gently from the other side. Sarah, startled yet relieved to hear a familiar voice, responded, 'Yes, I'm here, Emily. Just give me a moment, I'll open the door.'">>\\\\

\# Example 3: The text only satisfies part of the constraint.\\ 
- Main goal: Write a short story in which the protagonist meets an animal. \\
- Constraint: The walk should take place in a public space in a summer day. \\
- Text: As John strolled through the park one crisp autumn morning, he noticed the usual red and gold leaves blanketing the path. Today, however, a stray dog, thin and shivering, approached him. He hesitated, then offered his hand for the dog to sniff. It flinched at first, but soon warmed up to him. As they walked together, John wondered if he should take it home or find its owner. \\
- Your response: <<"answer": "Partially", "reasoning": "The text mentions that the character walks in a park, which satisifies the constraint that the setting is a public place. However, the walk takes place in an autumn morning, which violates the constraint that the walk takes place in a summer day", "quote": "As John strolled through the park one crisp autumn morning, he noticed the usual red and gold leaves blanketing the path...">>\\\\

\# Instruction \\
\#\# Main Goal\\
\{goal\}\\\\

\#\# Constraint\\
\{constraint\}\\\\

\# Text\\
\{text\}\\\\

DO NOT output anything else other than the response, which starts with "<<" and ending with ">>".\\\\
\# Your response

\\\bottomrule
\end{tabular}
\caption{\label{prompt:llm-eval} Prompt to evaluate whether a text follows a constraint or not. \{goal\}, \{constraint\}, and \{text\} are placeholders that will be replaced with actual content.}
\end{table*}

\newpage
\section{Generations from Mistral-Instruct, \data-SFT, and \data-ORPO}
We show generations from Mistral-Instruct, \data-SFT, and \data-ORPO in Table \ref{tab:result-qual}.

\begin{table*}[ht!]
    \centering
    \scriptsize
    \begin{tabular}{p{1.5cm}p{4.75cm}p{4.75cm}p{3.25cm}}
    \toprule
         \textbf{Constraint}&  \textbf{\data-I-ORPO}& \textbf{\data-SFT}& \textbf{Mistral-Instruct}\\
         \midrule
         Detail the onset of a serious, life-altering illness, including the initial diagnosis and the impact it had on the quality of life and daily activities of the person.& Being diagnosed with MSA often means a rapid decline, and it didn't take long for it to rob Elizabeth of her steady gait and coordination. There were numerous falls, steps hit, food dropped. There was depression, panic, and the wrenching witness to a vibrant, funny, independent woman being stripped of her abilities. Slowly, as she grew more debilitated, we made adjustments. She stopped driving, which was an excruciating blow to her. She then lost her ability to use her laptop. She struggled to get into the shower, and then used a shower chair. She used a wheelchair, and then an electric scooter. She became a master of calling an ambulance before her body decided to keep her upright.& She was diagnosed with advanced pulmonary hypertension in July of 2013, immediately after heart failure, which was the result of an array of intestinal diseases and surgeries that dated back to 2001...For several years, she was able to continue attending to her business and loved entertaining friends and partying at home. However, as her illness advanced, it became impossible for her to live the life that she loved. She was no longer able to play her beloved music, do needle point, bake, cook or garden. She struggled to even go for walks, even though she had many appointments at local hospitals and clinics. She became primarily housebound, with only short trips to the grocery store or to see a pet warden. She was reduced to having nearly 24hrs a day of assistance, administration of a nebulizer, and hospice nurses intervening if her symptoms increased.&But the seasons turned, and so did Amelia's fate. One fateful autumn day, she received the news that would forever change her life: a diagnosis of a degenerative neurological condition. The prognosis was grim, a cruel reminder that even the strongest among us are mortal. Her once nimble hands could no longer hold a pen to write her poetry, her once steady legs could no longer carry her to tend to her garden. Her light began to dim, and the darkness of her illness cast a long, tragic shadow over her life.\\
         \midrule
         Begin the narrative with a detailed description of the setting, emphasizing the time of day, the condition of the sea, and the immediate surroundings to establish a strong sense of place.&  The ship cut through the waves as evening fell. The sky was painted the colors of a summer sunset, deepening to pink, orange, purple, and green in the west as the sun sank toward the horizon. The air was warm, redolent with salt, and the ship rocked with the weight of the cargos that it carried in its hold. Ahead, the Pale Horse sped through the waves, its white sails snapping in the wind that billowed from the west. A flock of seabirds wheeling above the ship took no notice of its passage. Nearby, two smaller vessels followed, one with a low, curving bow that gave it an almost feminine grace in the breeze, the other a large merchant ship, weighed down with grain and other trade items. To the south east, the Pale Horse's sister ship, the Red Lion, was already out of sight. & The sun was going down, a pink glow still covering the horizon, but its fading light made the water look almost pitch black. A light surf broke against the stones of the harbor's seawall. High on the shore, where the promontory formed a horseshoe shape, there was a small port, and beyond that the main wall of the city, the tower of the Castello Giustinian, and the cathedral could be seen against the blackening sky. Here, off-shore, there were only the dark blue waters of the Adriatic. There was no color, no life, and the sea stretched away for miles. & The evening sun began its descent towards the western horizon, casting long shadows on the undulating waters of the River Tiber. The air, thick with the fragrant bouquet of jasmine and rose from the gardens of Palazzo Borgia, was tinged with a hint of chill as autumn's tender tendrils crept closer to the heart of the Italian Peninsula.\\
     \bottomrule
    \end{tabular}
    \caption{Example generations from \data-I-ORPO, \data-SFT, Mistral-Instruct. All three generations make a decent attempt at following the given constraint.}
    \label{tab:result-qual}
\end{table*}

\section{Comparable training setup for \data-SFT and \data-I-ORPO}
Here, we report the results for \data-SFT-single, where we fine-tune Mistral-Instruct-7B-v0.2 using the same instruction setup as I-ORPO (including only one constraint in the instruction). Table~\ref{tab:suri_comparison} demonstrates that \data-SFT-single underperforms \data-I-ORPO in all aspects. While \data-SFT-single does generate longer text compared to the baseline models, it exhibits more repetitions and lower ranking accuracy than both \data-I-ORPO and \data-SFT. Our qualitative analysis of 30 text samples generated by \data-SFT-single shows that the generations contain more gibberish, satisfy fewer constraints, and are generally harder to read. These findings reinforce our original claim that I-ORPO outperforms SFT.

\begin{table*}[htbp]
\small
\centering
\begin{tabular}{lcc}
\toprule
\textbf{Metric} & \textbf{Suri-SFT-Single} & \textbf{Suri-I-ORPO} \\
\midrule
Average number of tokens in test set generation & 4007 & 5102 \\
Number of test set generations with 5-gram repetitions & 32\% & 24\% \\
Number of test set generations with 10-gram repetitions & 4\% & 3\% \\
Ranking accuracy [$(M,M)$, $(M,M/2)$, $(M,1)$, $(M/2,M/2)$, $(1,1)$] & [99, 94, 65, 96, 76] & [100, 100, 98, 100, 98] \\
\bottomrule
\end{tabular}
\caption{Comparison of Suri-SFT-Single and Suri-I-ORPO performance metrics}
\label{tab:suri_comparison}
\end{table*}

\section{Fine-tuning on \data~does not significantly degrade performance in short-form instruction following tasks}
We measure the performance of \data-I-ORPO and \data-SFT on popular benchmarks using lm-evaluation-harness~\citep{eval-harness}. Our findings indicate that our fine-tuned models do not significantly degrade the performance of the baseline instruct model (Table \ref{tab:model_harness}). In fact, they slightly improve performance on most benchmarks, with the exceptions of HellaSwag, WinoGrande, and Arc-challenge.

\begin{table*}[htbp]
\scriptsize
\centering
\begin{tabular}{lcccccccc}
\toprule
\textbf{Model} & \textbf{MMLU} & \textbf{HellaSwag} & \textbf{WinoGrande} & \textbf{PIQA} & \textbf{Arc-e} & \textbf{Arc-c} & \textbf{GSM8K} & \textbf{GPQA} \\ 
\midrule
I-ORPO & 62.81\% & 80.31\% & 72.69\% & 80.63\% & 81.99\% & 53.67\% & 42.23\% & 31.25\% \\ 
SFT & 57.79\% & 80.21\% & 71.82\% & 81.18\% & 82.03\% & 53.07\% & 41.09\% & 30.25\% \\ 
Mistral-7b-instruct-v2 (baseline) & 58.73\% & 83.67\% & 73.48\% & 80.25\% & 81.36\% & 54.44\% & 42.08\% & 26.34\% \\ 
\bottomrule
\end{tabular}
\caption{Comparison of model performance across various tasks}
\label{tab:model_harness}
\end{table*}

\section{GPT-4o-mini's performance on \data}
\begin{table}[ht]
\centering
\begin{tabular}{lcc}
\toprule
\textbf{Model} & \textbf{GPT-4o-mini} & \textbf{Suri-I-ORPO} \\
\midrule
Average number of tokens in test set generation & 1134 & 5102 \\
Number of test set generations with 5-gram repetitions & 6\% & 24\% \\
Number of test set generations with 10-gram repetitions & 1\% & 3\% \\
\bottomrule
\end{tabular}
\caption{\label{tab:mini} Comparison between GPT-4o-mini and Suri-I-ORPO on test set generations.}
\end{table}

We note that although GPT-4o-mini produces less repetitive text, it can generate only an average of 1,134 tokens, which is still lower than Mixtral-8x7B-Instruct (Table \ref{tab:mini}). The higher repetition rate in the fine-tuned models may simply be due to these models generating longer text. Upon analyzing 30 generation samples from GPT-4o-mini, we observe that while the model can satisfy the constraints, it still suffers from formulaic generation and unnatural incorporation of those constraints.

%% file: main.bbl
\begin{thebibliography}{69}
\providecommand{\natexlab}[1]{#1}

\bibitem[{AI@Meta(2024)}]{llama3modelcard}
AI@Meta. 2024.
\newblock \href {https://github.com/meta-llama/llama3/blob/main/MODEL_CARD.md} {Llama 3 model card}.

\bibitem[{Anthropic(2024)}]{claude_sonnet}
Anthropic. 2024.
\newblock \href {https://www.anthropic.com/news/claude-3-5-sonnet} {Introducing {Claude} 3.5 {Sonnet}}.

\bibitem[{Askell et~al.(2021)Askell, Bai, Chen, Drain, Ganguli, Henighan, Jones, Joseph, Mann, DasSarma, Elhage, Hatfield-Dodds, Hernandez, Kernion, Ndousse, Olsson, Amodei, Brown, Clark, McCandlish, Olah, and Kaplan}]{askell_general_2021}
Amanda Askell, Yuntao Bai, Anna Chen, Dawn Drain, Deep Ganguli, Tom Henighan, Andy Jones, Nicholas Joseph, Ben Mann, Nova DasSarma, Nelson Elhage, Zac Hatfield-Dodds, Danny Hernandez, Jackson Kernion, Kamal Ndousse, Catherine Olsson, Dario Amodei, Tom Brown, Jack Clark, Sam McCandlish, Chris Olah, and Jared Kaplan. 2021.
\newblock \href {https://doi.org/10.48550/arXiv.2112.00861} {A {General} {Language} {Assistant} as a {Laboratory} for {Alignment}}.
\newblock \emph{arXiv preprint}.
\newblock ArXiv:2112.00861 [cs].

\bibitem[{Bai et~al.(2022)Bai, Kadavath, Kundu, Askell, Kernion, Jones, Chen, Goldie, Mirhoseini, McKinnon, Chen, Olsson, Olah, Hernandez, Drain, Ganguli, Li, Tran-Johnson, Perez, Kerr, Mueller, Ladish, Landau, Ndousse, Lukosuite, Lovitt, Sellitto, Elhage, Schiefer, Mercado, DasSarma, Lasenby, Larson, Ringer, Johnston, Kravec, Showk, Fort, Lanham, Telleen-Lawton, Conerly, Henighan, Hume, Bowman, Hatfield-Dodds, Mann, Amodei, Joseph, McCandlish, Brown, and Kaplan}]{bai2022constitutional}
Yuntao Bai, Saurav Kadavath, Sandipan Kundu, Amanda Askell, Jackson Kernion, Andy Jones, Anna Chen, Anna Goldie, Azalia Mirhoseini, Cameron McKinnon, Carol Chen, Catherine Olsson, Christopher Olah, Danny Hernandez, Dawn Drain, Deep Ganguli, Dustin Li, Eli Tran-Johnson, Ethan Perez, Jamie Kerr, Jared Mueller, Jeffrey Ladish, Joshua Landau, Kamal Ndousse, Kamile Lukosuite, Liane Lovitt, Michael Sellitto, Nelson Elhage, Nicholas Schiefer, Noemi Mercado, Nova DasSarma, Robert Lasenby, Robin Larson, Sam Ringer, Scott Johnston, Shauna Kravec, Sheer~El Showk, Stanislav Fort, Tamera Lanham, Timothy Telleen-Lawton, Tom Conerly, Tom Henighan, Tristan Hume, Samuel~R. Bowman, Zac Hatfield-Dodds, Ben Mann, Dario Amodei, Nicholas Joseph, Sam McCandlish, Tom Brown, and Jared Kaplan. 2022.
\newblock \href {https://arxiv.org/abs/2212.08073} {Constitutional ai: Harmlessness from ai feedback}.
\newblock \emph{Preprint}, arXiv:2212.08073.

\bibitem[{Bai et~al.(2024)Bai, Lv, Zhang, He, Qi, Hou, Tang, Dong, and Li}]{bai_longalign_2024}
Yushi Bai, Xin Lv, Jiajie Zhang, Yuze He, Ji~Qi, Lei Hou, Jie Tang, Yuxiao Dong, and Juanzi Li. 2024.
\newblock \href {http://arxiv.org/abs/2401.18058} {{LongAlign}: {A} {Recipe} for {Long} {Context} {Alignment} of {Large} {Language} {Models}!}
\newblock \emph{arXiv preprint}.
\newblock ArXiv:2401.18058 [cs].

\bibitem[{BernhardClemm(2023)}]{ercexpo_us_news_domains_2023}
BernhardClemm. 2023.
\newblock \href {https://doi.org/10.5281/zenodo.7651047} {{ercexpo/us-news-domains: v2.0.0 (v2.0.0)}}.

\bibitem[{Card et~al.(2020)Card, Henderson, Khandelwal, Jia, Mahowald, and Jurafsky}]{card-etal-2020-little}
Dallas Card, Peter Henderson, Urvashi Khandelwal, Robin Jia, Kyle Mahowald, and Dan Jurafsky. 2020.
\newblock \href {https://doi.org/10.18653/v1/2020.emnlp-main.745} {With little power comes great responsibility}.
\newblock In \emph{Proceedings of the 2020 Conference on Empirical Methods in Natural Language Processing (EMNLP)}, pages 9263--9274, Online. Association for Computational Linguistics.

\bibitem[{Chen et~al.(2024{\natexlab{a}})Chen, Malladi, Zhang, Chen, Zhang, Ranganath, and Cho}]{chen2024preference}
Angelica Chen, Sadhika Malladi, Lily~H. Zhang, Xinyi Chen, Qiuyi Zhang, Rajesh Ranganath, and Kyunghyun Cho. 2024{\natexlab{a}}.
\newblock \href {https://arxiv.org/abs/2405.19534} {Preference learning algorithms do not learn preference rankings}.
\newblock \emph{Preprint}, arXiv:2405.19534.

\bibitem[{Chen et~al.(2024{\natexlab{b}})Chen, Qian, Tang, Lai, Liu, Han, and Jia}]{chen2024longlora}
Yukang Chen, Shengju Qian, Haotian Tang, Xin Lai, Zhijian Liu, Song Han, and Jiaya Jia. 2024{\natexlab{b}}.
\newblock \href {https://arxiv.org/abs/2309.12307} {Longlora: Efficient fine-tuning of long-context large language models}.
\newblock \emph{Preprint}, arXiv:2309.12307.

\bibitem[{Chung et~al.(2022)Chung, Hou, Longpre, Zoph, Tay, Fedus, Li, Wang, Dehghani, Brahma, Webson, Gu, Dai, Suzgun, Chen, Chowdhery, Castro-Ros, Pellat, Robinson, Valter, Narang, Mishra, Yu, Zhao, Huang, Dai, Yu, Petrov, Chi, Dean, Devlin, Roberts, Zhou, Le, and Wei}]{chung2022scaling}
Hyung~Won Chung, Le~Hou, Shayne Longpre, Barret Zoph, Yi~Tay, William Fedus, Yunxuan Li, Xuezhi Wang, Mostafa Dehghani, Siddhartha Brahma, Albert Webson, Shixiang~Shane Gu, Zhuyun Dai, Mirac Suzgun, Xinyun Chen, Aakanksha Chowdhery, Alex Castro-Ros, Marie Pellat, Kevin Robinson, Dasha Valter, Sharan Narang, Gaurav Mishra, Adams Yu, Vincent Zhao, Yanping Huang, Andrew Dai, Hongkun Yu, Slav Petrov, Ed~H. Chi, Jeff Dean, Jacob Devlin, Adam Roberts, Denny Zhou, Quoc~V. Le, and Jason Wei. 2022.
\newblock \href {https://arxiv.org/abs/2210.11416} {Scaling instruction-finetuned language models}.
\newblock \emph{Preprint}, arXiv:2210.11416.

\bibitem[{Computer(2023)}]{together2023redpajama}
Together Computer. 2023.
\newblock \href {https://github.com/togethercomputer/RedPajama-Data} {Redpajama: an open dataset for training large language models}.

\bibitem[{Conover et~al.(2023)Conover, Hayes, Mathur, Xie, Wan, Shah, Ghodsi, Wendell, Zaharia, and Xin}]{DatabricksBlog2023DollyV2}
Mike Conover, Matt Hayes, Ankit Mathur, Jianwei Xie, Jun Wan, Sam Shah, Ali Ghodsi, Patrick Wendell, Matei Zaharia, and Reynold Xin. 2023.
\newblock \href {https://www.databricks.com/blog/2023/04/12/dolly-first-open-commercially-viable-instruction-tuned-llm} {Free dolly: Introducing the world's first truly open instruction-tuned llm}.

\bibitem[{Dao(2024)}]{dao2023flashattention2}
Tri Dao. 2024.
\newblock Flash{A}ttention-2: Faster attention with better parallelism and work partitioning.
\newblock In \emph{International Conference on Learning Representations (ICLR)}.

\bibitem[{Ethayarajh et~al.(2024)Ethayarajh, Xu, Muennighoff, Jurafsky, and Kiela}]{ethayarajh2024kto}
Kawin Ethayarajh, Winnie Xu, Niklas Muennighoff, Dan Jurafsky, and Douwe Kiela. 2024.
\newblock \href {https://arxiv.org/abs/2402.01306} {Kto: Model alignment as prospect theoretic optimization}.
\newblock \emph{Preprint}, arXiv:2402.01306.

\bibitem[{Fan et~al.(2018)Fan, Lewis, and Dauphin}]{fan2018hierarchical}
Angela Fan, Mike Lewis, and Yann Dauphin. 2018.
\newblock \href {https://arxiv.org/abs/1805.04833} {Hierarchical neural story generation}.
\newblock \emph{Preprint}, arXiv:1805.04833.

\bibitem[{Gao et~al.(2020)Gao, Biderman, Black, Golding, Hoppe, Foster, Phang, He, Thite, Nabeshima, Presser, and Leahy}]{pile}
Leo Gao, Stella Biderman, Sid Black, Laurence Golding, Travis Hoppe, Charles Foster, Jason Phang, Horace He, Anish Thite, Noa Nabeshima, Shawn Presser, and Connor Leahy. 2020.
\newblock The {P}ile: An 800gb dataset of diverse text for language modeling.
\newblock \emph{arXiv preprint arXiv:2101.00027}.

\bibitem[{Gao et~al.(2024)Gao, Tow, Abbasi, Biderman, Black, DiPofi, Foster, Golding, Hsu, Le~Noac'h, Li, McDonell, Muennighoff, Ociepa, Phang, Reynolds, Schoelkopf, Skowron, Sutawika, Tang, Thite, Wang, Wang, and Zou}]{eval-harness}
Leo Gao, Jonathan Tow, Baber Abbasi, Stella Biderman, Sid Black, Anthony DiPofi, Charles Foster, Laurence Golding, Jeffrey Hsu, Alain Le~Noac'h, Haonan Li, Kyle McDonell, Niklas Muennighoff, Chris Ociepa, Jason Phang, Laria Reynolds, Hailey Schoelkopf, Aviya Skowron, Lintang Sutawika, Eric Tang, Anish Thite, Ben Wang, Kevin Wang, and Andy Zou. 2024.
\newblock \href {https://doi.org/10.5281/zenodo.12608602} {A framework for few-shot language model evaluation}.

\bibitem[{Guan et~al.(2021)Guan, Mao, Fan, Liu, Ding, and Huang}]{guan2021long}
Jian Guan, Xiaoxi Mao, Changjie Fan, Zitao Liu, Wenbiao Ding, and Minlie Huang. 2021.
\newblock \href {https://arxiv.org/abs/2105.08963} {Long text generation by modeling sentence-level and discourse-level coherence}.
\newblock \emph{Preprint}, arXiv:2105.08963.

\bibitem[{Hong et~al.(2024)Hong, Lee, and Thorne}]{hong2024orpo}
Jiwoo Hong, Noah Lee, and James Thorne. 2024.
\newblock \href {https://arxiv.org/abs/2403.07691} {Orpo: Monolithic preference optimization without reference model}.
\newblock \emph{Preprint}, arXiv:2403.07691.

\bibitem[{Hu et~al.(2021)Hu, Shen, Wallis, Allen-Zhu, Li, Wang, Wang, and Chen}]{hu2021lora}
Edward~J. Hu, Yelong Shen, Phillip Wallis, Zeyuan Allen-Zhu, Yuanzhi Li, Shean Wang, Lu~Wang, and Weizhu Chen. 2021.
\newblock \href {https://arxiv.org/abs/2106.09685} {Lora: Low-rank adaptation of large language models}.
\newblock \emph{Preprint}, arXiv:2106.09685.

\bibitem[{Jiang et~al.(2023)Jiang, Sablayrolles, Mensch, Bamford, Chaplot, de~las Casas, Bressand, Lengyel, Lample, Saulnier, Lavaud, Lachaux, Stock, Scao, Lavril, Wang, Lacroix, and Sayed}]{jiang2023mistral}
Albert~Q. Jiang, Alexandre Sablayrolles, Arthur Mensch, Chris Bamford, Devendra~Singh Chaplot, Diego de~las Casas, Florian Bressand, Gianna Lengyel, Guillaume Lample, Lucile Saulnier, Lélio~Renard Lavaud, Marie-Anne Lachaux, Pierre Stock, Teven~Le Scao, Thibaut Lavril, Thomas Wang, Timothée Lacroix, and William~El Sayed. 2023.
\newblock \href {https://arxiv.org/abs/2310.06825} {Mistral 7b}.
\newblock \emph{Preprint}, arXiv:2310.06825.

\bibitem[{Jiang et~al.(2024)Jiang, Sablayrolles, Roux, Mensch, Savary, Bamford, Chaplot, de~las Casas, Hanna, Bressand, Lengyel, Bour, Lample, Lavaud, Saulnier, Lachaux, Stock, Subramanian, Yang, Antoniak, Scao, Gervet, Lavril, Wang, Lacroix, and Sayed}]{jiang2024mixtral}
Albert~Q. Jiang, Alexandre Sablayrolles, Antoine Roux, Arthur Mensch, Blanche Savary, Chris Bamford, Devendra~Singh Chaplot, Diego de~las Casas, Emma~Bou Hanna, Florian Bressand, Gianna Lengyel, Guillaume Bour, Guillaume Lample, Lélio~Renard Lavaud, Lucile Saulnier, Marie-Anne Lachaux, Pierre Stock, Sandeep Subramanian, Sophia Yang, Szymon Antoniak, Teven~Le Scao, Théophile Gervet, Thibaut Lavril, Thomas Wang, Timothée Lacroix, and William~El Sayed. 2024.
\newblock \href {https://arxiv.org/abs/2401.04088} {Mixtral of experts}.
\newblock \emph{Preprint}, arXiv:2401.04088.

\bibitem[{Kim et~al.(2024)Kim, Chang, Karpinska, Garimella, Manjunatha, Lo, Goyal, and Iyyer}]{kim2024fables}
Yekyung Kim, Yapei Chang, Marzena Karpinska, Aparna Garimella, Varun Manjunatha, Kyle Lo, Tanya Goyal, and Mohit Iyyer. 2024.
\newblock \href {https://arxiv.org/abs/2404.01261} {Fables: Evaluating faithfulness and content selection in book-length summarization}.
\newblock \emph{Preprint}, arXiv:2404.01261.

\bibitem[{Kreutzer et~al.(2018)Kreutzer, Uyheng, and Riezler}]{kreutzer-etal-2018-reliability}
Julia Kreutzer, Joshua Uyheng, and Stefan Riezler. 2018.
\newblock \href {https://doi.org/10.18653/v1/P18-1165} {Reliability and learnability of human bandit feedback for sequence-to-sequence reinforcement learning}.
\newblock In \emph{Proceedings of the 56th Annual Meeting of the Association for Computational Linguistics (Volume 1: Long Papers)}, pages 1777--1788, Melbourne, Australia. Association for Computational Linguistics.

\bibitem[{Kwon et~al.(2023)Kwon, Li, Zhuang, Sheng, Zheng, Yu, Gonzalez, Zhang, and Stoica}]{kwon2023efficient}
Woosuk Kwon, Zhuohan Li, Siyuan Zhuang, Ying Sheng, Lianmin Zheng, Cody~Hao Yu, Joseph~E. Gonzalez, Hao Zhang, and Ion Stoica. 2023.
\newblock Efficient memory management for large language model serving with pagedattention.
\newblock In \emph{Proceedings of the ACM SIGOPS 29th Symposium on Operating Systems Principles}.

\bibitem[{Köksal et~al.(2023)Köksal, Schick, Korhonen, and Schütze}]{koksal_longform_2023}
Abdullatif Köksal, Timo Schick, Anna Korhonen, and Hinrich Schütze. 2023.
\newblock \href {http://arxiv.org/abs/2304.08460} {{LongForm}: {Optimizing} {Instruction} {Tuning} for {Long} {Text} {Generation} with {Corpus} {Extraction}}.
\newblock \emph{arXiv preprint}.
\newblock ArXiv:2304.08460 [cs].

\bibitem[{Köpf et~al.(2023)Köpf, Kilcher, von Rütte, Anagnostidis, Tam, Stevens, Barhoum, Duc, Stanley, Nagyfi, ES, Suri, Glushkov, Dantuluri, Maguire, Schuhmann, Nguyen, and Mattick}]{köpf2023openassistant}
Andreas Köpf, Yannic Kilcher, Dimitri von Rütte, Sotiris Anagnostidis, Zhi-Rui Tam, Keith Stevens, Abdullah Barhoum, Nguyen~Minh Duc, Oliver Stanley, Richárd Nagyfi, Shahul ES, Sameer Suri, David Glushkov, Arnav Dantuluri, Andrew Maguire, Christoph Schuhmann, Huu Nguyen, and Alexander Mattick. 2023.
\newblock \href {https://arxiv.org/abs/2304.07327} {Openassistant conversations -- democratizing large language model alignment}.
\newblock \emph{Preprint}, arXiv:2304.07327.

\bibitem[{Lee et~al.(2023)Lee, Phatale, Mansoor, Lu, Mesnard, Bishop, Carbune, and Rastogi}]{lee2023rlaif}
Harrison Lee, Samrat Phatale, Hassan Mansoor, Kellie Lu, Thomas Mesnard, Colton Bishop, Victor Carbune, and Abhinav Rastogi. 2023.
\newblock Rlaif: Scaling reinforcement learning from human feedback with ai feedback.
\newblock \emph{arXiv preprint arXiv:2309.00267}.

\bibitem[{Li et~al.(2016)Li, Galley, Brockett, Gao, and Dolan}]{li-etal-2016-diversity}
Jiwei Li, Michel Galley, Chris Brockett, Jianfeng Gao, and Bill Dolan. 2016.
\newblock \href {https://doi.org/10.18653/v1/N16-1014} {A diversity-promoting objective function for neural conversation models}.
\newblock In \emph{Proceedings of the 2016 Conference of the North {A}merican Chapter of the Association for Computational Linguistics: Human Language Technologies}, pages 110--119, San Diego, California. Association for Computational Linguistics.

\bibitem[{Li et~al.(2023)Li, Yu, Zhou, Schick, Zettlemoyer, Levy, Weston, and Lewis}]{li2023self}
Xian Li, Ping Yu, Chunting Zhou, Timo Schick, Luke Zettlemoyer, Omer Levy, Jason Weston, and Mike Lewis. 2023.
\newblock Self-alignment with instruction backtranslation.
\newblock \emph{arXiv preprint arXiv:2308.06259}.

\bibitem[{Liu et~al.(2023)Liu, Sferrazza, and Abbeel}]{liu2023chain}
Hao Liu, Carmelo Sferrazza, and Pieter Abbeel. 2023.
\newblock \href {https://arxiv.org/abs/2302.02676} {Chain of hindsight aligns language models with feedback}.
\newblock \emph{Preprint}, arXiv:2302.02676.

\bibitem[{Malaviya et~al.(2024)Malaviya, Agrawal, Ganchev, Srinivasan, Huot, Berant, Yatskar, Das, Lapata, and Alberti}]{malaviya2024dolomites}
Chaitanya Malaviya, Priyanka Agrawal, Kuzman Ganchev, Pranesh Srinivasan, Fantine Huot, Jonathan Berant, Mark Yatskar, Dipanjan Das, Mirella Lapata, and Chris Alberti. 2024.
\newblock Dolomites: Domain-specific long-form methodical tasks.
\newblock In \emph{arXiv}.

\bibitem[{Mangrulkar et~al.(2022)Mangrulkar, Gugger, Debut, Belkada, Paul, and Bossan}]{peft}
Sourab Mangrulkar, Sylvain Gugger, Lysandre Debut, Younes Belkada, Sayak Paul, and Benjamin Bossan. 2022.
\newblock Peft: State-of-the-art parameter-efficient fine-tuning methods.
\newblock \url{https://github.com/huggingface/peft}.

\bibitem[{Mishra et~al.(2022)Mishra, Khashabi, Baral, and Hajishirzi}]{mishra2022crosstask}
Swaroop Mishra, Daniel Khashabi, Chitta Baral, and Hannaneh Hajishirzi. 2022.
\newblock \href {https://arxiv.org/abs/2104.08773} {Cross-task generalization via natural language crowdsourcing instructions}.
\newblock \emph{Preprint}, arXiv:2104.08773.

\bibitem[{Mostafazadeh et~al.(2016)Mostafazadeh, Chambers, He, Parikh, Batra, Vanderwende, Kohli, and Allen}]{mostafazadeh-etal-2016-corpus}
Nasrin Mostafazadeh, Nathanael Chambers, Xiaodong He, Devi Parikh, Dhruv Batra, Lucy Vanderwende, Pushmeet Kohli, and James Allen. 2016.
\newblock \href {https://doi.org/10.18653/v1/N16-1098} {A corpus and cloze evaluation for deeper understanding of commonsense stories}.
\newblock In \emph{Proceedings of the 2016 Conference of the North {A}merican Chapter of the Association for Computational Linguistics: Human Language Technologies}, pages 839--849, San Diego, California. Association for Computational Linguistics.

\bibitem[{{OpenAI}(2024)}]{openai-gpt4o}
{OpenAI}. 2024.
\newblock \href {https://openai.com/index/hello-gpt-4o/} {{Model release blog: GPT-4o}}.
\newblock Technical report, OpenAI.
\newblock Accessed: 2024-05-23.

\bibitem[{Ouyang et~al.(2022)Ouyang, Wu, Jiang, Almeida, Wainwright, Mishkin, Zhang, Agarwal, Slama, Ray, Schulman, Hilton, Kelton, Miller, Simens, Askell, Welinder, Christiano, Leike, and Lowe}]{ouyang_training_2022}
Long Ouyang, Jeff Wu, Xu~Jiang, Diogo Almeida, Carroll~L. Wainwright, Pamela Mishkin, Chong Zhang, Sandhini Agarwal, Katarina Slama, Alex Ray, John Schulman, Jacob Hilton, Fraser Kelton, Luke Miller, Maddie Simens, Amanda Askell, Peter Welinder, Paul Christiano, Jan Leike, and Ryan Lowe. 2022.
\newblock \href {https://doi.org/10.48550/arXiv.2203.02155} {Training language models to follow instructions with human feedback}.
\newblock \emph{arXiv preprint}.
\newblock ArXiv:2203.02155 [cs].

\bibitem[{Presser(2020)}]{presser}
Shawn Presser. 2020.
\newblock \href {https://twitter.com/theshawwn/status/1320282149329784833} {Books3}.

\bibitem[{Rafailov et~al.(2023)Rafailov, Sharma, Mitchell, Ermon, Manning, and Finn}]{rafailov2023direct}
Rafael Rafailov, Archit Sharma, Eric Mitchell, Stefano Ermon, Christopher~D. Manning, and Chelsea Finn. 2023.
\newblock \href {https://arxiv.org/abs/2305.18290} {Direct preference optimization: Your language model is secretly a reward model}.
\newblock \emph{Preprint}, arXiv:2305.18290.

\bibitem[{Rajani et~al.(2023)Rajani, Tunstall, Beeching, Lambert, Rush, and Wolf}]{no_robots}
Nazneen Rajani, Lewis Tunstall, Edward Beeching, Nathan Lambert, Alexander~M. Rush, and Thomas Wolf. 2023.
\newblock No robots.
\newblock \url{https://huggingface.co/datasets/HuggingFaceH4/no_robots}.

\bibitem[{Ramamurthy et~al.(2023)Ramamurthy, Ammanabrolu, Brantley, Hessel, Sifa, Bauckhage, Hajishirzi, and Choi}]{ramamurthy2023reinforcement}
Rajkumar Ramamurthy, Prithviraj Ammanabrolu, Kianté Brantley, Jack Hessel, Rafet Sifa, Christian Bauckhage, Hannaneh Hajishirzi, and Yejin Choi. 2023.
\newblock \href {https://arxiv.org/abs/2210.01241} {Is reinforcement learning (not) for natural language processing: Benchmarks, baselines, and building blocks for natural language policy optimization}.
\newblock \emph{Preprint}, arXiv:2210.01241.

\bibitem[{Rasley et~al.(2020)Rasley, Rajbhandari, Ruwase, and He}]{deepspeed}
Jeff Rasley, Samyam Rajbhandari, Olatunji Ruwase, and Yuxiong He. 2020.
\newblock \href {https://doi.org/10.1145/3394486.3406703} {Deepspeed: System optimizations enable training deep learning models with over 100 billion parameters}.
\newblock In \emph{Proceedings of the 26th ACM SIGKDD International Conference on Knowledge Discovery \& Data Mining}, KDD '20, page 3505–3506, New York, NY, USA. Association for Computing Machinery.

\bibitem[{Sanh et~al.(2022)Sanh, Webson, Raffel, Bach, Sutawika, Alyafeai, Chaffin, Stiegler, Scao, Raja, Dey, Bari, Xu, Thakker, Sharma, Szczechla, Kim, Chhablani, Nayak, Datta, Chang, Jiang, Wang, Manica, Shen, Yong, Pandey, Bawden, Wang, Neeraj, Rozen, Sharma, Santilli, Fevry, Fries, Teehan, Bers, Biderman, Gao, Wolf, and Rush}]{sanh2022multitask}
Victor Sanh, Albert Webson, Colin Raffel, Stephen~H. Bach, Lintang Sutawika, Zaid Alyafeai, Antoine Chaffin, Arnaud Stiegler, Teven~Le Scao, Arun Raja, Manan Dey, M~Saiful Bari, Canwen Xu, Urmish Thakker, Shanya~Sharma Sharma, Eliza Szczechla, Taewoon Kim, Gunjan Chhablani, Nihal Nayak, Debajyoti Datta, Jonathan Chang, Mike Tian-Jian Jiang, Han Wang, Matteo Manica, Sheng Shen, Zheng~Xin Yong, Harshit Pandey, Rachel Bawden, Thomas Wang, Trishala Neeraj, Jos Rozen, Abheesht Sharma, Andrea Santilli, Thibault Fevry, Jason~Alan Fries, Ryan Teehan, Tali Bers, Stella Biderman, Leo Gao, Thomas Wolf, and Alexander~M. Rush. 2022.
\newblock \href {https://arxiv.org/abs/2110.08207} {Multitask prompted training enables zero-shot task generalization}.
\newblock \emph{Preprint}, arXiv:2110.08207.

\bibitem[{See et~al.(2019)See, Pappu, Saxena, Yerukola, and Manning}]{see2019massively}
Abigail See, Aneesh Pappu, Rohun Saxena, Akhila Yerukola, and Christopher~D. Manning. 2019.
\newblock \href {https://arxiv.org/abs/1909.10705} {Do massively pretrained language models make better storytellers?}
\newblock \emph{Preprint}, arXiv:1909.10705.

\bibitem[{Shaham et~al.(2022)Shaham, Segal, Ivgi, Efrat, Yoran, Haviv, Gupta, Xiong, Geva, Berant, and Levy}]{shaham2022scrolls}
Uri Shaham, Elad Segal, Maor Ivgi, Avia Efrat, Ori Yoran, Adi Haviv, Ankit Gupta, Wenhan Xiong, Mor Geva, Jonathan Berant, and Omer Levy. 2022.
\newblock \href {https://arxiv.org/abs/2201.03533} {Scrolls: Standardized comparison over long language sequences}.
\newblock \emph{Preprint}, arXiv:2201.03533.

\bibitem[{Stiennon et~al.(2022)Stiennon, Ouyang, Wu, Ziegler, Lowe, Voss, Radford, Amodei, and Christiano}]{stiennon2022learning}
Nisan Stiennon, Long Ouyang, Jeff Wu, Daniel~M. Ziegler, Ryan Lowe, Chelsea Voss, Alec Radford, Dario Amodei, and Paul Christiano. 2022.
\newblock \href {https://arxiv.org/abs/2009.01325} {Learning to summarize from human feedback}.
\newblock \emph{Preprint}, arXiv:2009.01325.

\bibitem[{Sun et~al.(2022)Sun, Thai, and Iyyer}]{sun2022chapterbreak}
Simeng Sun, Katherine Thai, and Mohit Iyyer. 2022.
\newblock Chapterbreak: A challenge dataset for long-range language models.
\newblock \emph{arXiv preprint arXiv:2204.10878}.

\bibitem[{Taori et~al.(2023)Taori, Gulrajani, Zhang, Dubois, Li, Guestrin, Liang, and Hashimoto}]{alpaca}
Rohan Taori, Ishaan Gulrajani, Tianyi Zhang, Yann Dubois, Xuechen Li, Carlos Guestrin, Percy Liang, and Tatsunori~B. Hashimoto. 2023.
\newblock Stanford alpaca: An instruction-following llama model.
\newblock \url{https://github.com/tatsu-lab/stanford_alpaca}.

\bibitem[{Team et~al.(2024)Team, Georgiev, Lei, Burnell, Bai, Gulati, Tanzer, Vincent, Pan, Wang, Mariooryad, Ding, Geng, Alcober, Frostig, Omernick, Walker, Paduraru, Sorokin, Tacchetti, Gaffney, Daruki, Sercinoglu, Gleicher, Love, Voigtlaender, Jain, Surita, Mohamed, Blevins, Ahn, Zhu, Kawintiranon, Firat, Gu, Zhang, Rahtz, Faruqui, Clay, Gilmer, Co-Reyes, Penchev, Zhu, Morioka, Hui, Haridasan, Campos, Mahdieh, Guo, Hassan, Kilgour, Vezer, Cheng, de~Liedekerke, Goyal, Barham, Strouse, Noury, Adler, Sundararajan, Vikram, Lepikhin, Paganini, Garcia, Yang, Valter, Trebacz, Vodrahalli, Asawaroengchai, Ring, Kalb, Soares, Brahma, Steiner, Yu, Mentzer, He, Gonzalez, Xu, Kaufman, Shafey, Oh, Hennigan, van~den Driessche, Odoom, Lucic, Roelofs, Lall, Marathe, Chan, Ontanon, He, Teplyashin, Lai, Crone, Damoc, Ho, Riedel, Lenc, Yeh, Chowdhery, Xu, Kazemi, Amid, Petrushkina, Swersky, Khodaei, Chen, Larkin, Pinto, Yan, Badia, Patil, Hansen, Orr, Arnold, Grimstad, Dai, Douglas, Sinha, Yadav, Chen, Gribovskaya, Austin,
  Zhao, Patel, Komarek, Austin, Borgeaud, Friso, Goyal, Caine, Cao, Chung, Lamm, Barth-Maron, Kagohara, Olszewska, Chen, Shivakumar, Agarwal, Godhia, Rajwar, Snaider, Dotiwalla, Liu, Barua, Ungureanu, Zhang, Batsaikhan, Wirth, Qin, Danihelka, Doshi, Chadwick, Chen, Jain, Le, Kar, Gurumurthy, Li, Sang, Liu, Lamprou, Munoz, Lintz, Mehta, Howard, Reynolds, Aroyo, Wang, Blanco, Cassirer, Griffith, Das, Lee, Sygnowski, Fisher, Besley, Powell, Ahmed, Paulus, Reitter, Borsos, Joshi, Pope, Hand, Selo, Jain, Sethi, Goel, Makino, May, Yang, Schalkwyk, Butterfield, Hauth, Goldin, Hawkins, Senter, Brin, Woodman, Ritter, Noland, Giang, Bolina, Lee, Blyth, Mackinnon, Reid, Sarvana, Silver, Chen, Wang, Maggiore, Chang, Attaluri, Thornton, Chiu, Bunyan, Levine, Chung, Eltyshev, Si, Lillicrap, Brady, Aggarwal, Wu, Xu, McIlroy, Badola, Sandhu, Moreira, Stokowiec, Hemsley, Li, Tudor, Shyam, Rahimtoroghi, Haykal, Sprechmann, Zhou, Mincu, Li, Addanki, Krishna, Wu, Frechette, Eyal, Dafoe, Lacey, Whang, Avrahami, Zhang, Taropa,
  Lin, Toyama, Rutherford, Sano, Choe, Tomala, Safranek-Shrader, Kassner, Pajarskas, Harvey, Sechrist, Fortunato, Lyu, Elsayed, Kuang, Lottes, Chu, Jia, Chen, Humphreys, Baumli, Tao, Samuel, dos Santos, Andreassen, Rakićević, Grewe, Kumar, Winkler, Caton, Brock, Dalmia, Sheahan, Barr, Miao, Natsev, Devlin, Behbahani, Prost, Sun, Myaskovsky, Pillai, Hurt, Lazaridou, Xiong, Zheng, Pardo, Li, Horgan, Stanton, Ambar, Xia, Lince, Wang, Mustafa, Webson, Lee, Anil, Wicke, Dozat, Sinha, Piqueras, Dabir, Upadhyay, Boral, Hendricks, Fry, Djolonga, Su, Walker, Labanowski, Huang, Misra, Chen, Skerry-Ryan, Singh, Rijhwani, Yu, Castro-Ros, Changpinyo, Datta, Bagri, Hrafnkelsson, Maggioni, Zheng, Sulsky, Hou, Paine, Yang, Riesa, Rogozinska, Marcus, Badawy, Zhang, Wang, Miller, Greer, Sjos, Nova, Zen, Chaabouni, Rosca, Jiang, Chen, Liu, Sainath, Krikun, Polozov, Lespiau, Newlan, Cankara, Kwak, Xu, Chen, Coenen, Meyer, Tsihlas, Ma, Gottweis, Xing, Gu, Miao, Frank, Cankara, Ganapathy, Dasgupta, Hughes-Fitt, Chen, Reid, Rong,
  Fan, van Amersfoort, Zhuang, Cohen, Gu, Mohananey, Ilic, Tobin, Wieting, Bortsova, Thacker, Wang, Caveness, Chiu, Sezener, Kaskasoli, Baker, Millican, Elhawaty, Aisopos, Lebsack, Byrd, Dai, Jia, Wiethoff, Davoodi, Weston, Yagati, Ahuja, Gao, Pundak, Zhang, Azzam, Sim, Caelles, Keeling, Sharma, Swing, Li, Liu, Bostock, Bansal, Nado, Anand, Lipschultz, Karmarkar, Proleev, Ittycheriah, Yeganeh, Polovets, Faust, Sun, Rrustemi, Li, Shivanna, Liu, Welty, Lebron, Baddepudi, Krause, Parisotto, Soricut, Xu, Bloxwich, Johnson, Neyshabur, Mao-Jones, Wang, Ramasesh, Abbas, Guez, Segal, Nguyen, Svensson, Hou, York, Milan, Bridgers, Gworek, Tagliasacchi, Lee-Thorp, Chang, Guseynov, Hartman, Kwong, Zhao, Kashem, Cole, Miech, Tanburn, Phuong, Pavetic, Cevey, Comanescu, Ives, Yang, Du, Li, Zhang, Iinuma, Hu, Roy, Bijwadia, Zhu, Martins, Saputro, Gergely, Zheng, Jia, Antonoglou, Sadovsky, Gu, Bi, Andreev, Samangooei, Khan, Kocisky, Filos, Kumar, Bishop, Yu, Hodkinson, Mittal, Shah, Moufarek, Cheng, Bloniarz, Lee, Pejman,
  Michel, Spencer, Feinberg, Xiong, Savinov, Smith, Shakeri, Tran, Chesus, Bohnet, Tucker, von Glehn, Muir, Mao, Kazawa, Slone, Soparkar, Shrivastava, Cobon-Kerr, Sharman, Pavagadhi, Araya, Misiunas, Ghelani, Laskin, Barker, Li, Briukhov, Houlsby, Glaese, Lakshminarayanan, Schucher, Tang, Collins, Lim, Feng, Recasens, Lai, Magni, Cao, Siddhant, Ashwood, Orbay, Dehghani, Brennan, He, Xu, Gao, Saroufim, Molloy, Wu, Arnold, Chang, Schrittwieser, Buchatskaya, Radpour, Polacek, Giordano, Bapna, Tokumine, Hellendoorn, Sottiaux, Cogan, Severyn, Saleh, Thakoor, Shefey, Qiao, Gaba, yiin Chang, Swanson, Zhang, Lee, Rubenstein, Song, Kwiatkowski, Koop, Kannan, Kao, Schuh, Stjerngren, Ghiasi, Gibson, Vilnis, Yuan, Ferreira, Kamath, Klimenko, Franko, Xiao, Bhattacharya, Patel, Wang, Morris, Strudel, Sharma, Choy, Hashemi, Landon, Finkelstein, Jhakra, Frye, Barnes, Mauger, Daun, Baatarsukh, Tung, Farhan, Michalewski, Viola, de~Chaumont~Quitry, Lan, Hudson, Wang, Fischer, Zheng, White, Dragan, baptiste Alayrac, Ni, Pritzel,
  Iwanicki, Isard, Bulanova, Zilka, Dyer, Sachan, Srinivasan, Muckenhirn, Cai, Mandhane, Tariq, Rae, Wang, Ayoub, FitzGerald, Zhao, Han, Alberti, Garrette, Krishnakumar, Gimenez, Levskaya, Sohn, Matak, Iturrate, Chang, Xiang, Cao, Ranka, Brown, Hutter, Mirrokni, Chen, Yao, Egyed, Galilee, Liechty, Kallakuri, Palmer, Ghemawat, Liu, Tao, Thornton, Green, Jasarevic, Lin, Cotruta, Tan, Fiedel, Yu, Chi, Neitz, Heitkaemper, Sinha, Zhou, Sun, Kaed, Hulse, Mishra, Georgaki, Kudugunta, Farabet, Shafran, Vlasic, Tsitsulin, Ananthanarayanan, Carin, Su, Sun, V, Carvajal, Broder, Comsa, Repina, Wong, Chen, Hawkins, Filonov, Loher, Hirnschall, Wang, Ye, Burns, Cate, Wright, Piccinini, Zhang, Lin, Gog, Kulizhskaya, Sreevatsa, Song, Cobo, Iyer, Tekur, Garrido, Xiao, Kemp, Zheng, Li, Agarwal, Ngani, Goshvadi, Santamaria-Fernandez, Fica, Chen, Gorgolewski, Sun, Garg, Ye, Eslami, Hua, Simon, Joshi, Kim, Tenney, Potluri, Thiet, Yuan, Luisier, Chronopoulou, Scellato, Srinivasan, Chen, Koverkathu, Dalibard, Xu, Saeta, Anderson,
  Sellam, Fernando, Huot, Jung, Varadarajan, Quinn, Raul, Le, Habalov, Clark, Jalan, Bullard, Singhal, Luong, Wang, Rajayogam, Eisenschlos, Jia, Finchelstein, Yakubovich, Balle, Fink, Agarwal, Li, Dvijotham, Pal, Kang, Konzelmann, Beattie, Dousse, Wu, Crocker, Elkind, Jonnalagadda, Lee, Holtmann-Rice, Kallarackal, Liu, Vnukov, Vats, Invernizzi, Jafari, Zhou, Taylor, Prendki, Wu, Eccles, Liu, Kopparapu, Beaufays, Angermueller, Marzoca, Sarcar, Dib, Stanway, Perbet, Trdin, Sterneck, Khorlin, Li, Wu, Goenka, Madras, Goldshtein, Gierke, Zhou, Liu, Liang, White, Li, Singh, Bahargam, Epstein, Basu, Lao, Ozturel, Crous, Zhai, Lu, Tung, Gaur, Walton, Dixon, Zhang, Globerson, Uy, Bolt, Wiles, Nasr, Shumailov, Selvi, Piccinno, Aguilar, McCarthy, Khalman, Shukla, Galic, Carpenter, Villela, Zhang, Richardson, Martens, Bosnjak, Belle, Seibert, Alnahlawi, McWilliams, Singh, Louis, Ding, Popovici, Simicich, Knight, Mehta, Gupta, Shi, Fatehi, Mitrovic, Grills, Pagadora, Petrova, Eisenbud, Zhang, Yates, Mittal, Tripuraneni,
  Assael, Brovelli, Jain, Velimirovic, Akbulut, Mu, Macherey, Kumar, Xu, Qureshi, Comanici, Wiesner, Gong, Ruddock, Bauer, Felt, GP, Arnab, Zelle, Rothfuss, Rosgen, Shenoy, Seybold, Li, Mudigonda, Erdogan, Xia, Simsa, Michi, Yao, Yew, Kan, Caswell, Radebaugh, Elisseeff, Valenzuela, McKinney, Paterson, Cui, Latorre-Chimoto, Kim, Zeng, Durden, Ponnapalli, Sosea, Choquette-Choo, Manyika, Robenek, Vashisht, Pereira, Lam, Velic, Owusu-Afriyie, Lee, Bolukbasi, Parrish, Lu, Park, Venkatraman, Talbert, Rosique, Cheng, Sozanschi, Paszke, Kumar, Austin, Li, Salama, Kim, Dukkipati, Baryshnikov, Kaplanis, Sheng, Chervonyi, Unlu, de~Las~Casas, Askham, Tunyasuvunakool, Gimeno, Poder, Kwak, Miecnikowski, Mirrokni, Dimitriev, Parisi, Liu, Tsai, Shevlane, Kouridi, Garmon, Goedeckemeyer, Brown, Vijayakumar, Elqursh, Jazayeri, Huang, Carthy, Hoover, Kim, Kumar, Chen, Biles, Bingham, Rosen, Wang, Tan, Engel, Pongetti, de~Cesare, Hwang, Yu, Pullman, Narayanan, Levin, Gopal, Li, Aharoni, Trinh, Lo, Casagrande, Vij, Matthey,
  Ramadhana, Matthews, Carey, Johnson, Goranova, Shah, Ashraf, Dasgupta, Larsen, Wang, Vuyyuru, Jiang, Ijazi, Osawa, Smith, Boppana, Bilal, Koizumi, Xu, Altun, Shabat, Bariach, Korchemniy, Choo, Ronneberger, Iwuanyanwu, Zhao, Soergel, Hsieh, Cai, Iqbal, Sundermeyer, Chen, Bursztein, Malaviya, Biadsy, Shroff, Dhillon, Latkar, Dyer, Forbes, Nicosia, Nikolaev, Greene, Georgiev, Wang, Martin, Sedghi, Zhang, Banzal, Fritz, Rao, Wang, Zhang, Patraucean, Du, Mordatch, Jurin, Liu, Dubey, Mohan, Nowakowski, Ion, Wei, Tojo, Raad, Hudson, Keshava, Agrawal, Ramirez, Wu, Nguyen, Liu, Sewak, Petrini, Choi, Philips, Wang, Bica, Garg, Wilkiewicz, Agrawal, Li, Guo, Xue, Shaik, Leach, Khan, Wiesinger, Jerome, Chakladar, Wang, Ornduff, Abu, Ghaffarkhah, Wainwright, Cortes, Liu, Maynez, Terzis, Samangouei, Mansour, Kępa, Aubet, Algymr, Banica, Weisz, Orban, Senges, Andrejczuk, Geller, Santo, Anklin, Merey, Baeuml, Strohman, Bai, Petrov, Wu, Hassabis, Kavukcuoglu, Dean, and Vinyals}]{geminiteam2024gemini15unlockingmultimodal}
Gemini Team, Petko Georgiev, Ving~Ian Lei, Ryan Burnell, Libin Bai, Anmol Gulati, Garrett Tanzer, Damien Vincent, Zhufeng Pan, Shibo Wang, Soroosh Mariooryad, Yifan Ding, Xinyang Geng, Fred Alcober, Roy Frostig, Mark Omernick, Lexi Walker, Cosmin Paduraru, Christina Sorokin, Andrea Tacchetti, Colin Gaffney, Samira Daruki, Olcan Sercinoglu, Zach Gleicher, Juliette Love, Paul Voigtlaender, Rohan Jain, Gabriela Surita, Kareem Mohamed, Rory Blevins, Junwhan Ahn, Tao Zhu, Kornraphop Kawintiranon, Orhan Firat, Yiming Gu, Yujing Zhang, Matthew Rahtz, Manaal Faruqui, Natalie Clay, Justin Gilmer, JD~Co-Reyes, Ivo Penchev, Rui Zhu, Nobuyuki Morioka, Kevin Hui, Krishna Haridasan, Victor Campos, Mahdis Mahdieh, Mandy Guo, Samer Hassan, Kevin Kilgour, Arpi Vezer, Heng-Tze Cheng, Raoul de~Liedekerke, Siddharth Goyal, Paul Barham, DJ~Strouse, Seb Noury, Jonas Adler, Mukund Sundararajan, Sharad Vikram, Dmitry Lepikhin, Michela Paganini, Xavier Garcia, Fan Yang, Dasha Valter, Maja Trebacz, Kiran Vodrahalli, Chulayuth
  Asawaroengchai, Roman Ring, Norbert Kalb, Livio~Baldini Soares, Siddhartha Brahma, David Steiner, Tianhe Yu, Fabian Mentzer, Antoine He, Lucas Gonzalez, Bibo Xu, Raphael~Lopez Kaufman, Laurent~El Shafey, Junhyuk Oh, Tom Hennigan, George van~den Driessche, Seth Odoom, Mario Lucic, Becca Roelofs, Sid Lall, Amit Marathe, Betty Chan, Santiago Ontanon, Luheng He, Denis Teplyashin, Jonathan Lai, Phil Crone, Bogdan Damoc, Lewis Ho, Sebastian Riedel, Karel Lenc, Chih-Kuan Yeh, Aakanksha Chowdhery, Yang Xu, Mehran Kazemi, Ehsan Amid, Anastasia Petrushkina, Kevin Swersky, Ali Khodaei, Gowoon Chen, Chris Larkin, Mario Pinto, Geng Yan, Adria~Puigdomenech Badia, Piyush Patil, Steven Hansen, Dave Orr, Sebastien M.~R. Arnold, Jordan Grimstad, Andrew Dai, Sholto Douglas, Rishika Sinha, Vikas Yadav, Xi~Chen, Elena Gribovskaya, Jacob Austin, Jeffrey Zhao, Kaushal Patel, Paul Komarek, Sophia Austin, Sebastian Borgeaud, Linda Friso, Abhimanyu Goyal, Ben Caine, Kris Cao, Da-Woon Chung, Matthew Lamm, Gabe Barth-Maron, Thais
  Kagohara, Kate Olszewska, Mia Chen, Kaushik Shivakumar, Rishabh Agarwal, Harshal Godhia, Ravi Rajwar, Javier Snaider, Xerxes Dotiwalla, Yuan Liu, Aditya Barua, Victor Ungureanu, Yuan Zhang, Bat-Orgil Batsaikhan, Mateo Wirth, James Qin, Ivo Danihelka, Tulsee Doshi, Martin Chadwick, Jilin Chen, Sanil Jain, Quoc Le, Arjun Kar, Madhu Gurumurthy, Cheng Li, Ruoxin Sang, Fangyu Liu, Lampros Lamprou, Rich Munoz, Nathan Lintz, Harsh Mehta, Heidi Howard, Malcolm Reynolds, Lora Aroyo, Quan Wang, Lorenzo Blanco, Albin Cassirer, Jordan Griffith, Dipanjan Das, Stephan Lee, Jakub Sygnowski, Zach Fisher, James Besley, Richard Powell, Zafarali Ahmed, Dominik Paulus, David Reitter, Zalan Borsos, Rishabh Joshi, Aedan Pope, Steven Hand, Vittorio Selo, Vihan Jain, Nikhil Sethi, Megha Goel, Takaki Makino, Rhys May, Zhen Yang, Johan Schalkwyk, Christina Butterfield, Anja Hauth, Alex Goldin, Will Hawkins, Evan Senter, Sergey Brin, Oliver Woodman, Marvin Ritter, Eric Noland, Minh Giang, Vijay Bolina, Lisa Lee, Tim Blyth, Ian
  Mackinnon, Machel Reid, Obaid Sarvana, David Silver, Alexander Chen, Lily Wang, Loren Maggiore, Oscar Chang, Nithya Attaluri, Gregory Thornton, Chung-Cheng Chiu, Oskar Bunyan, Nir Levine, Timothy Chung, Evgenii Eltyshev, Xiance Si, Timothy Lillicrap, Demetra Brady, Vaibhav Aggarwal, Boxi Wu, Yuanzhong Xu, Ross McIlroy, Kartikeya Badola, Paramjit Sandhu, Erica Moreira, Wojciech Stokowiec, Ross Hemsley, Dong Li, Alex Tudor, Pranav Shyam, Elahe Rahimtoroghi, Salem Haykal, Pablo Sprechmann, Xiang Zhou, Diana Mincu, Yujia Li, Ravi Addanki, Kalpesh Krishna, Xiao Wu, Alexandre Frechette, Matan Eyal, Allan Dafoe, Dave Lacey, Jay Whang, Thi Avrahami, Ye~Zhang, Emanuel Taropa, Hanzhao Lin, Daniel Toyama, Eliza Rutherford, Motoki Sano, HyunJeong Choe, Alex Tomala, Chalence Safranek-Shrader, Nora Kassner, Mantas Pajarskas, Matt Harvey, Sean Sechrist, Meire Fortunato, Christina Lyu, Gamaleldin Elsayed, Chenkai Kuang, James Lottes, Eric Chu, Chao Jia, Chih-Wei Chen, Peter Humphreys, Kate Baumli, Connie Tao, Rajkumar
  Samuel, Cicero~Nogueira dos Santos, Anders Andreassen, Nemanja Rakićević, Dominik Grewe, Aviral Kumar, Stephanie Winkler, Jonathan Caton, Andrew Brock, Sid Dalmia, Hannah Sheahan, Iain Barr, Yingjie Miao, Paul Natsev, Jacob Devlin, Feryal Behbahani, Flavien Prost, Yanhua Sun, Artiom Myaskovsky, Thanumalayan~Sankaranarayana Pillai, Dan Hurt, Angeliki Lazaridou, Xi~Xiong, Ce~Zheng, Fabio Pardo, Xiaowei Li, Dan Horgan, Joe Stanton, Moran Ambar, Fei Xia, Alejandro Lince, Mingqiu Wang, Basil Mustafa, Albert Webson, Hyo Lee, Rohan Anil, Martin Wicke, Timothy Dozat, Abhishek Sinha, Enrique Piqueras, Elahe Dabir, Shyam Upadhyay, Anudhyan Boral, Lisa~Anne Hendricks, Corey Fry, Josip Djolonga, Yi~Su, Jake Walker, Jane Labanowski, Ronny Huang, Vedant Misra, Jeremy Chen, RJ~Skerry-Ryan, Avi Singh, Shruti Rijhwani, Dian Yu, Alex Castro-Ros, Beer Changpinyo, Romina Datta, Sumit Bagri, Arnar~Mar Hrafnkelsson, Marcello Maggioni, Daniel Zheng, Yury Sulsky, Shaobo Hou, Tom~Le Paine, Antoine Yang, Jason Riesa, Dominika
  Rogozinska, Dror Marcus, Dalia~El Badawy, Qiao Zhang, Luyu Wang, Helen Miller, Jeremy Greer, Lars~Lowe Sjos, Azade Nova, Heiga Zen, Rahma Chaabouni, Mihaela Rosca, Jiepu Jiang, Charlie Chen, Ruibo Liu, Tara Sainath, Maxim Krikun, Alex Polozov, Jean-Baptiste Lespiau, Josh Newlan, Zeyncep Cankara, Soo Kwak, Yunhan Xu, Phil Chen, Andy Coenen, Clemens Meyer, Katerina Tsihlas, Ada Ma, Juraj Gottweis, Jinwei Xing, Chenjie Gu, Jin Miao, Christian Frank, Zeynep Cankara, Sanjay Ganapathy, Ishita Dasgupta, Steph Hughes-Fitt, Heng Chen, David Reid, Keran Rong, Hongmin Fan, Joost van Amersfoort, Vincent Zhuang, Aaron Cohen, Shixiang~Shane Gu, Anhad Mohananey, Anastasija Ilic, Taylor Tobin, John Wieting, Anna Bortsova, Phoebe Thacker, Emma Wang, Emily Caveness, Justin Chiu, Eren Sezener, Alex Kaskasoli, Steven Baker, Katie Millican, Mohamed Elhawaty, Kostas Aisopos, Carl Lebsack, Nathan Byrd, Hanjun Dai, Wenhao Jia, Matthew Wiethoff, Elnaz Davoodi, Albert Weston, Lakshman Yagati, Arun Ahuja, Isabel Gao, Golan Pundak,
  Susan Zhang, Michael Azzam, Khe~Chai Sim, Sergi Caelles, James Keeling, Abhanshu Sharma, Andy Swing, YaGuang Li, Chenxi Liu, Carrie~Grimes Bostock, Yamini Bansal, Zachary Nado, Ankesh Anand, Josh Lipschultz, Abhijit Karmarkar, Lev Proleev, Abe Ittycheriah, Soheil~Hassas Yeganeh, George Polovets, Aleksandra Faust, Jiao Sun, Alban Rrustemi, Pen Li, Rakesh Shivanna, Jeremiah Liu, Chris Welty, Federico Lebron, Anirudh Baddepudi, Sebastian Krause, Emilio Parisotto, Radu Soricut, Zheng Xu, Dawn Bloxwich, Melvin Johnson, Behnam Neyshabur, Justin Mao-Jones, Renshen Wang, Vinay Ramasesh, Zaheer Abbas, Arthur Guez, Constant Segal, Duc~Dung Nguyen, James Svensson, Le~Hou, Sarah York, Kieran Milan, Sophie Bridgers, Wiktor Gworek, Marco Tagliasacchi, James Lee-Thorp, Michael Chang, Alexey Guseynov, Ale~Jakse Hartman, Michael Kwong, Ruizhe Zhao, Sheleem Kashem, Elizabeth Cole, Antoine Miech, Richard Tanburn, Mary Phuong, Filip Pavetic, Sebastien Cevey, Ramona Comanescu, Richard Ives, Sherry Yang, Cosmo Du, Bo~Li, Zizhao
  Zhang, Mariko Iinuma, Clara~Huiyi Hu, Aurko Roy, Shaan Bijwadia, Zhenkai Zhu, Danilo Martins, Rachel Saputro, Anita Gergely, Steven Zheng, Dawei Jia, Ioannis Antonoglou, Adam Sadovsky, Shane Gu, Yingying Bi, Alek Andreev, Sina Samangooei, Mina Khan, Tomas Kocisky, Angelos Filos, Chintu Kumar, Colton Bishop, Adams Yu, Sarah Hodkinson, Sid Mittal, Premal Shah, Alexandre Moufarek, Yong Cheng, Adam Bloniarz, Jaehoon Lee, Pedram Pejman, Paul Michel, Stephen Spencer, Vladimir Feinberg, Xuehan Xiong, Nikolay Savinov, Charlotte Smith, Siamak Shakeri, Dustin Tran, Mary Chesus, Bernd Bohnet, George Tucker, Tamara von Glehn, Carrie Muir, Yiran Mao, Hideto Kazawa, Ambrose Slone, Kedar Soparkar, Disha Shrivastava, James Cobon-Kerr, Michael Sharman, Jay Pavagadhi, Carlos Araya, Karolis Misiunas, Nimesh Ghelani, Michael Laskin, David Barker, Qiujia Li, Anton Briukhov, Neil Houlsby, Mia Glaese, Balaji Lakshminarayanan, Nathan Schucher, Yunhao Tang, Eli Collins, Hyeontaek Lim, Fangxiaoyu Feng, Adria Recasens, Guangda Lai,
  Alberto Magni, Nicola~De Cao, Aditya Siddhant, Zoe Ashwood, Jordi Orbay, Mostafa Dehghani, Jenny Brennan, Yifan He, Kelvin Xu, Yang Gao, Carl Saroufim, James Molloy, Xinyi Wu, Seb Arnold, Solomon Chang, Julian Schrittwieser, Elena Buchatskaya, Soroush Radpour, Martin Polacek, Skye Giordano, Ankur Bapna, Simon Tokumine, Vincent Hellendoorn, Thibault Sottiaux, Sarah Cogan, Aliaksei Severyn, Mohammad Saleh, Shantanu Thakoor, Laurent Shefey, Siyuan Qiao, Meenu Gaba, Shuo yiin Chang, Craig Swanson, Biao Zhang, Benjamin Lee, Paul~Kishan Rubenstein, Gan Song, Tom Kwiatkowski, Anna Koop, Ajay Kannan, David Kao, Parker Schuh, Axel Stjerngren, Golnaz Ghiasi, Gena Gibson, Luke Vilnis, Ye~Yuan, Felipe~Tiengo Ferreira, Aishwarya Kamath, Ted Klimenko, Ken Franko, Kefan Xiao, Indro Bhattacharya, Miteyan Patel, Rui Wang, Alex Morris, Robin Strudel, Vivek Sharma, Peter Choy, Sayed~Hadi Hashemi, Jessica Landon, Mara Finkelstein, Priya Jhakra, Justin Frye, Megan Barnes, Matthew Mauger, Dennis Daun, Khuslen Baatarsukh, Matthew
  Tung, Wael Farhan, Henryk Michalewski, Fabio Viola, Felix de~Chaumont~Quitry, Charline~Le Lan, Tom Hudson, Qingze Wang, Felix Fischer, Ivy Zheng, Elspeth White, Anca Dragan, Jean baptiste Alayrac, Eric Ni, Alexander Pritzel, Adam Iwanicki, Michael Isard, Anna Bulanova, Lukas Zilka, Ethan Dyer, Devendra Sachan, Srivatsan Srinivasan, Hannah Muckenhirn, Honglong Cai, Amol Mandhane, Mukarram Tariq, Jack~W. Rae, Gary Wang, Kareem Ayoub, Nicholas FitzGerald, Yao Zhao, Woohyun Han, Chris Alberti, Dan Garrette, Kashyap Krishnakumar, Mai Gimenez, Anselm Levskaya, Daniel Sohn, Josip Matak, Inaki Iturrate, Michael~B. Chang, Jackie Xiang, Yuan Cao, Nishant Ranka, Geoff Brown, Adrian Hutter, Vahab Mirrokni, Nanxin Chen, Kaisheng Yao, Zoltan Egyed, Francois Galilee, Tyler Liechty, Praveen Kallakuri, Evan Palmer, Sanjay Ghemawat, Jasmine Liu, David Tao, Chloe Thornton, Tim Green, Mimi Jasarevic, Sharon Lin, Victor Cotruta, Yi-Xuan Tan, Noah Fiedel, Hongkun Yu, Ed~Chi, Alexander Neitz, Jens Heitkaemper, Anu Sinha, Denny
  Zhou, Yi~Sun, Charbel Kaed, Brice Hulse, Swaroop Mishra, Maria Georgaki, Sneha Kudugunta, Clement Farabet, Izhak Shafran, Daniel Vlasic, Anton Tsitsulin, Rajagopal Ananthanarayanan, Alen Carin, Guolong Su, Pei Sun, Shashank V, Gabriel Carvajal, Josef Broder, Iulia Comsa, Alena Repina, William Wong, Warren~Weilun Chen, Peter Hawkins, Egor Filonov, Lucia Loher, Christoph Hirnschall, Weiyi Wang, Jingchen Ye, Andrea Burns, Hardie Cate, Diana~Gage Wright, Federico Piccinini, Lei Zhang, Chu-Cheng Lin, Ionel Gog, Yana Kulizhskaya, Ashwin Sreevatsa, Shuang Song, Luis~C. Cobo, Anand Iyer, Chetan Tekur, Guillermo Garrido, Zhuyun Xiao, Rupert Kemp, Huaixiu~Steven Zheng, Hui Li, Ananth Agarwal, Christel Ngani, Kati Goshvadi, Rebeca Santamaria-Fernandez, Wojciech Fica, Xinyun Chen, Chris Gorgolewski, Sean Sun, Roopal Garg, Xinyu Ye, S.~M.~Ali Eslami, Nan Hua, Jon Simon, Pratik Joshi, Yelin Kim, Ian Tenney, Sahitya Potluri, Lam~Nguyen Thiet, Quan Yuan, Florian Luisier, Alexandra Chronopoulou, Salvatore Scellato, Praveen
  Srinivasan, Minmin Chen, Vinod Koverkathu, Valentin Dalibard, Yaming Xu, Brennan Saeta, Keith Anderson, Thibault Sellam, Nick Fernando, Fantine Huot, Junehyuk Jung, Mani Varadarajan, Michael Quinn, Amit Raul, Maigo Le, Ruslan Habalov, Jon Clark, Komal Jalan, Kalesha Bullard, Achintya Singhal, Thang Luong, Boyu Wang, Sujeevan Rajayogam, Julian Eisenschlos, Johnson Jia, Daniel Finchelstein, Alex Yakubovich, Daniel Balle, Michael Fink, Sameer Agarwal, Jing Li, Dj~Dvijotham, Shalini Pal, Kai Kang, Jaclyn Konzelmann, Jennifer Beattie, Olivier Dousse, Diane Wu, Remi Crocker, Chen Elkind, Siddhartha~Reddy Jonnalagadda, Jong Lee, Dan Holtmann-Rice, Krystal Kallarackal, Rosanne Liu, Denis Vnukov, Neera Vats, Luca Invernizzi, Mohsen Jafari, Huanjie Zhou, Lilly Taylor, Jennifer Prendki, Marcus Wu, Tom Eccles, Tianqi Liu, Kavya Kopparapu, Francoise Beaufays, Christof Angermueller, Andreea Marzoca, Shourya Sarcar, Hilal Dib, Jeff Stanway, Frank Perbet, Nejc Trdin, Rachel Sterneck, Andrey Khorlin, Dinghua Li, Xihui Wu,
  Sonam Goenka, David Madras, Sasha Goldshtein, Willi Gierke, Tong Zhou, Yaxin Liu, Yannie Liang, Anais White, Yunjie Li, Shreya Singh, Sanaz Bahargam, Mark Epstein, Sujoy Basu, Li~Lao, Adnan Ozturel, Carl Crous, Alex Zhai, Han Lu, Zora Tung, Neeraj Gaur, Alanna Walton, Lucas Dixon, Ming Zhang, Amir Globerson, Grant Uy, Andrew Bolt, Olivia Wiles, Milad Nasr, Ilia Shumailov, Marco Selvi, Francesco Piccinno, Ricardo Aguilar, Sara McCarthy, Misha Khalman, Mrinal Shukla, Vlado Galic, John Carpenter, Kevin Villela, Haibin Zhang, Harry Richardson, James Martens, Matko Bosnjak, Shreyas~Rammohan Belle, Jeff Seibert, Mahmoud Alnahlawi, Brian McWilliams, Sankalp Singh, Annie Louis, Wen Ding, Dan Popovici, Lenin Simicich, Laura Knight, Pulkit Mehta, Nishesh Gupta, Chongyang Shi, Saaber Fatehi, Jovana Mitrovic, Alex Grills, Joseph Pagadora, Dessie Petrova, Danielle Eisenbud, Zhishuai Zhang, Damion Yates, Bhavishya Mittal, Nilesh Tripuraneni, Yannis Assael, Thomas Brovelli, Prateek Jain, Mihajlo Velimirovic, Canfer
  Akbulut, Jiaqi Mu, Wolfgang Macherey, Ravin Kumar, Jun Xu, Haroon Qureshi, Gheorghe Comanici, Jeremy Wiesner, Zhitao Gong, Anton Ruddock, Matthias Bauer, Nick Felt, Anirudh GP, Anurag Arnab, Dustin Zelle, Jonas Rothfuss, Bill Rosgen, Ashish Shenoy, Bryan Seybold, Xinjian Li, Jayaram Mudigonda, Goker Erdogan, Jiawei Xia, Jiri Simsa, Andrea Michi, Yi~Yao, Christopher Yew, Steven Kan, Isaac Caswell, Carey Radebaugh, Andre Elisseeff, Pedro Valenzuela, Kay McKinney, Kim Paterson, Albert Cui, Eri Latorre-Chimoto, Solomon Kim, William Zeng, Ken Durden, Priya Ponnapalli, Tiberiu Sosea, Christopher~A. Choquette-Choo, James Manyika, Brona Robenek, Harsha Vashisht, Sebastien Pereira, Hoi Lam, Marko Velic, Denese Owusu-Afriyie, Katherine Lee, Tolga Bolukbasi, Alicia Parrish, Shawn Lu, Jane Park, Balaji Venkatraman, Alice Talbert, Lambert Rosique, Yuchung Cheng, Andrei Sozanschi, Adam Paszke, Praveen Kumar, Jessica Austin, Lu~Li, Khalid Salama, Wooyeol Kim, Nandita Dukkipati, Anthony Baryshnikov, Christos Kaplanis,
  XiangHai Sheng, Yuri Chervonyi, Caglar Unlu, Diego de~Las~Casas, Harry Askham, Kathryn Tunyasuvunakool, Felix Gimeno, Siim Poder, Chester Kwak, Matt Miecnikowski, Vahab Mirrokni, Alek Dimitriev, Aaron Parisi, Dangyi Liu, Tomy Tsai, Toby Shevlane, Christina Kouridi, Drew Garmon, Adrian Goedeckemeyer, Adam~R. Brown, Anitha Vijayakumar, Ali Elqursh, Sadegh Jazayeri, Jin Huang, Sara~Mc Carthy, Jay Hoover, Lucy Kim, Sandeep Kumar, Wei Chen, Courtney Biles, Garrett Bingham, Evan Rosen, Lisa Wang, Qijun Tan, David Engel, Francesco Pongetti, Dario de~Cesare, Dongseong Hwang, Lily Yu, Jennifer Pullman, Srini Narayanan, Kyle Levin, Siddharth Gopal, Megan Li, Asaf Aharoni, Trieu Trinh, Jessica Lo, Norman Casagrande, Roopali Vij, Loic Matthey, Bramandia Ramadhana, Austin Matthews, CJ~Carey, Matthew Johnson, Kremena Goranova, Rohin Shah, Shereen Ashraf, Kingshuk Dasgupta, Rasmus Larsen, Yicheng Wang, Manish~Reddy Vuyyuru, Chong Jiang, Joana Ijazi, Kazuki Osawa, Celine Smith, Ramya~Sree Boppana, Taylan Bilal, Yuma
  Koizumi, Ying Xu, Yasemin Altun, Nir Shabat, Ben Bariach, Alex Korchemniy, Kiam Choo, Olaf Ronneberger, Chimezie Iwuanyanwu, Shubin Zhao, David Soergel, Cho-Jui Hsieh, Irene Cai, Shariq Iqbal, Martin Sundermeyer, Zhe Chen, Elie Bursztein, Chaitanya Malaviya, Fadi Biadsy, Prakash Shroff, Inderjit Dhillon, Tejasi Latkar, Chris Dyer, Hannah Forbes, Massimo Nicosia, Vitaly Nikolaev, Somer Greene, Marin Georgiev, Pidong Wang, Nina Martin, Hanie Sedghi, John Zhang, Praseem Banzal, Doug Fritz, Vikram Rao, Xuezhi Wang, Jiageng Zhang, Viorica Patraucean, Dayou Du, Igor Mordatch, Ivan Jurin, Lewis Liu, Ayush Dubey, Abhi Mohan, Janek Nowakowski, Vlad-Doru Ion, Nan Wei, Reiko Tojo, Maria~Abi Raad, Drew~A. Hudson, Vaishakh Keshava, Shubham Agrawal, Kevin Ramirez, Zhichun Wu, Hoang Nguyen, Ji~Liu, Madhavi Sewak, Bryce Petrini, DongHyun Choi, Ivan Philips, Ziyue Wang, Ioana Bica, Ankush Garg, Jarek Wilkiewicz, Priyanka Agrawal, Xiaowei Li, Danhao Guo, Emily Xue, Naseer Shaik, Andrew Leach, Sadh~MNM Khan, Julia Wiesinger,
  Sammy Jerome, Abhishek Chakladar, Alek~Wenjiao Wang, Tina Ornduff, Folake Abu, Alireza Ghaffarkhah, Marcus Wainwright, Mario Cortes, Frederick Liu, Joshua Maynez, Andreas Terzis, Pouya Samangouei, Riham Mansour, Tomasz Kępa, François-Xavier Aubet, Anton Algymr, Dan Banica, Agoston Weisz, Andras Orban, Alexandre Senges, Ewa Andrejczuk, Mark Geller, Niccolo~Dal Santo, Valentin Anklin, Majd~Al Merey, Martin Baeuml, Trevor Strohman, Junwen Bai, Slav Petrov, Yonghui Wu, Demis Hassabis, Koray Kavukcuoglu, Jeffrey Dean, and Oriol Vinyals. 2024.
\newblock \href {https://arxiv.org/abs/2403.05530} {Gemini 1.5: Unlocking multimodal understanding across millions of tokens of context}.
\newblock \emph{Preprint}, arXiv:2403.05530.

\bibitem[{{The Association of Religion Data Archives}(2023)}]{ARDA_religiondictionary}
{The Association of Religion Data Archives}. 2023.
\newblock Religion dictionary.
\newblock \url{https://www.thearda.com/research/religion-dictionary}.
\newblock Accessed: 2024/01/15.

\bibitem[{Touvron et~al.(2023)Touvron, Martin, Stone, Albert, Almahairi, Babaei, Bashlykov, Batra, Bhargava, Bhosale et~al.}]{touvron2023llama}
Hugo Touvron, Louis Martin, Kevin Stone, Peter Albert, Amjad Almahairi, Yasmine Babaei, Nikolay Bashlykov, Soumya Batra, Prajjwal Bhargava, Shruti Bhosale, et~al. 2023.
\newblock Llama 2: Open foundation and fine-tuned chat models.
\newblock \emph{arXiv preprint arXiv:2307.09288}.

\bibitem[{Tunstall et~al.(2023)Tunstall, Beeching, Lambert, Rajani, Huang, Rasul, Rush, and Wolf}]{alignment_handbook2023}
Lewis Tunstall, Edward Beeching, Nathan Lambert, Nazneen Rajani, Shengyi Huang, Kashif Rasul, Alexander~M. Rush, and Thomas Wolf. 2023.
\newblock The alignment handbook.
\newblock \url{https://github.com/huggingface/alignment-handbook}.

\bibitem[{von Werra et~al.(2020)von Werra, Belkada, Tunstall, Beeching, Thrush, Lambert, and Huang}]{vonwerra2022trl}
Leandro von Werra, Younes Belkada, Lewis Tunstall, Edward Beeching, Tristan Thrush, Nathan Lambert, and Shengyi Huang. 2020.
\newblock Trl: Transformer reinforcement learning.
\newblock \url{https://github.com/huggingface/trl}.

\bibitem[{Wang et~al.(2023{\natexlab{a}})Wang, Durmus, Goodman, and Hashimoto}]{wang2023language}
Rose~E Wang, Esin Durmus, Noah Goodman, and Tatsunori Hashimoto. 2023{\natexlab{a}}.
\newblock \href {https://arxiv.org/abs/2203.11370} {Language modeling via stochastic processes}.
\newblock \emph{Preprint}, arXiv:2203.11370.

\bibitem[{Wang et~al.(2023{\natexlab{b}})Wang, Ivison, Dasigi, Hessel, Khot, Chandu, Wadden, MacMillan, Smith, Beltagy, and Hajishirzi}]{wang2023far}
Yizhong Wang, Hamish Ivison, Pradeep Dasigi, Jack Hessel, Tushar Khot, Khyathi~Raghavi Chandu, David Wadden, Kelsey MacMillan, Noah~A. Smith, Iz~Beltagy, and Hannaneh Hajishirzi. 2023{\natexlab{b}}.
\newblock \href {https://arxiv.org/abs/2306.04751} {How far can camels go? exploring the state of instruction tuning on open resources}.
\newblock \emph{Preprint}, arXiv:2306.04751.

\bibitem[{Wang et~al.(2023{\natexlab{c}})Wang, Kordi, Mishra, Liu, Smith, Khashabi, and Hajishirzi}]{wang2023selfinstruct}
Yizhong Wang, Yeganeh Kordi, Swaroop Mishra, Alisa Liu, Noah~A. Smith, Daniel Khashabi, and Hannaneh Hajishirzi. 2023{\natexlab{c}}.
\newblock \href {https://arxiv.org/abs/2212.10560} {Self-instruct: Aligning language models with self-generated instructions}.
\newblock \emph{Preprint}, arXiv:2212.10560.

\bibitem[{Wang et~al.(2022)Wang, Mishra, Alipoormolabashi, Kordi, Mirzaei, Arunkumar, Ashok, Dhanasekaran, Naik, Stap, Pathak, Karamanolakis, Lai, Purohit, Mondal, Anderson, Kuznia, Doshi, Patel, Pal, Moradshahi, Parmar, Purohit, Varshney, Kaza, Verma, Puri, Karia, Sampat, Doshi, Mishra, Reddy, Patro, Dixit, Shen, Baral, Choi, Smith, Hajishirzi, and Khashabi}]{wang2022supernaturalinstructions}
Yizhong Wang, Swaroop Mishra, Pegah Alipoormolabashi, Yeganeh Kordi, Amirreza Mirzaei, Anjana Arunkumar, Arjun Ashok, Arut~Selvan Dhanasekaran, Atharva Naik, David Stap, Eshaan Pathak, Giannis Karamanolakis, Haizhi~Gary Lai, Ishan Purohit, Ishani Mondal, Jacob Anderson, Kirby Kuznia, Krima Doshi, Maitreya Patel, Kuntal~Kumar Pal, Mehrad Moradshahi, Mihir Parmar, Mirali Purohit, Neeraj Varshney, Phani~Rohitha Kaza, Pulkit Verma, Ravsehaj~Singh Puri, Rushang Karia, Shailaja~Keyur Sampat, Savan Doshi, Siddhartha Mishra, Sujan Reddy, Sumanta Patro, Tanay Dixit, Xudong Shen, Chitta Baral, Yejin Choi, Noah~A. Smith, Hannaneh Hajishirzi, and Daniel Khashabi. 2022.
\newblock \href {https://arxiv.org/abs/2204.07705} {Super-naturalinstructions: Generalization via declarative instructions on 1600+ nlp tasks}.
\newblock \emph{Preprint}, arXiv:2204.07705.

\bibitem[{Wei et~al.(2021)Wei, Bosma, Zhao, Guu, Yu, Lester, Du, Dai, and Le}]{wei2021finetuned}
Jason Wei, Maarten Bosma, Vincent~Y Zhao, Kelvin Guu, Adams~Wei Yu, Brian Lester, Nan Du, Andrew~M Dai, and Quoc~V Le. 2021.
\newblock Finetuned language models are zero-shot learners.
\newblock \emph{arXiv preprint arXiv:2109.01652}.

\bibitem[{Wei et~al.(2022)Wei, Bosma, Zhao, Guu, Yu, Lester, Du, Dai, and Le}]{wei2022finetuned}
Jason Wei, Maarten Bosma, Vincent~Y. Zhao, Kelvin Guu, Adams~Wei Yu, Brian Lester, Nan Du, Andrew~M. Dai, and Quoc~V. Le. 2022.
\newblock \href {https://arxiv.org/abs/2109.01652} {Finetuned language models are zero-shot learners}.
\newblock \emph{Preprint}, arXiv:2109.01652.

\bibitem[{Wolf et~al.(2020)Wolf, Debut, Sanh, Chaumond, Delangue, Moi, Cistac, Rault, Louf, Funtowicz, Davison, Shleifer, von Platen, Ma, Jernite, Plu, Xu, Scao, Gugger, Drame, Lhoest, and Rush}]{wolf-etal-2020-transformers}
Thomas Wolf, Lysandre Debut, Victor Sanh, Julien Chaumond, Clement Delangue, Anthony Moi, Pierric Cistac, Tim Rault, Rémi Louf, Morgan Funtowicz, Joe Davison, Sam Shleifer, Patrick von Platen, Clara Ma, Yacine Jernite, Julien Plu, Canwen Xu, Teven~Le Scao, Sylvain Gugger, Mariama Drame, Quentin Lhoest, and Alexander~M. Rush. 2020.
\newblock \href {https://www.aclweb.org/anthology/2020.emnlp-demos.6} {Transformers: State-of-the-art natural language processing}.
\newblock In \emph{Proceedings of the 2020 Conference on Empirical Methods in Natural Language Processing: System Demonstrations}, pages 38--45, Online. Association for Computational Linguistics.

\bibitem[{Xiong et~al.(2023)Xiong, Liu, Molybog, Zhang, Bhargava, Hou, Martin, Rungta, Sankararaman, Oguz, Khabsa, Fang, Mehdad, Narang, Malik, Fan, Bhosale, Edunov, Lewis, Wang, and Ma}]{xiong_effective_2023}
Wenhan Xiong, Jingyu Liu, Igor Molybog, Hejia Zhang, Prajjwal Bhargava, Rui Hou, Louis Martin, Rashi Rungta, Karthik~Abinav Sankararaman, Barlas Oguz, Madian Khabsa, Han Fang, Yashar Mehdad, Sharan Narang, Kshitiz Malik, Angela Fan, Shruti Bhosale, Sergey Edunov, Mike Lewis, Sinong Wang, and Hao Ma. 2023.
\newblock \href {http://arxiv.org/abs/2309.16039} {Effective {Long}-{Context} {Scaling} of {Foundation} {Models}}.
\newblock \emph{arXiv preprint}.
\newblock ArXiv:2309.16039 [cs].

\bibitem[{Xu et~al.(2023{\natexlab{a}})Xu, Sun, Zheng, Geng, Zhao, Feng, Tao, and Jiang}]{xu2023wizardlm}
Can Xu, Qingfeng Sun, Kai Zheng, Xiubo Geng, Pu~Zhao, Jiazhan Feng, Chongyang Tao, and Daxin Jiang. 2023{\natexlab{a}}.
\newblock \href {https://arxiv.org/abs/2304.12244} {Wizardlm: Empowering large language models to follow complex instructions}.
\newblock \emph{Preprint}, arXiv:2304.12244.

\bibitem[{Xu et~al.(2023{\natexlab{b}})Xu, Guo, Duan, and McAuley}]{xu2023baize}
Canwen Xu, Daya Guo, Nan Duan, and Julian McAuley. 2023{\natexlab{b}}.
\newblock Baize: An open-source chat model with parameter-efficient tuning on self-chat data.
\newblock \emph{arXiv preprint arXiv:2304.01196}.

\bibitem[{Xu et~al.(2024)Xu, Lo, Soldaini, Kuehl, Choi, and Wadden}]{xu_kiwi_2024}
Fangyuan Xu, Kyle Lo, Luca Soldaini, Bailey Kuehl, Eunsol Choi, and David Wadden. 2024.
\newblock \href {http://arxiv.org/abs/2403.03866} {{KIWI}: {A} {Dataset} of {Knowledge}-{Intensive} {Writing} {Instructions} for {Answering} {Research} {Questions}}.
\newblock \emph{arXiv preprint}.
\newblock ArXiv:2403.03866 [cs].

\bibitem[{Xu et~al.(2023{\natexlab{c}})Xu, Song, Iyyer, and Choi}]{xu2023critical}
Fangyuan Xu, Yixiao Song, Mohit Iyyer, and Eunsol Choi. 2023{\natexlab{c}}.
\newblock \href {https://arxiv.org/abs/2305.18201} {A critical evaluation of evaluations for long-form question answering}.
\newblock \emph{Preprint}, arXiv:2305.18201.

\bibitem[{Yin et~al.(2023)Yin, Liu, Yin, Zhong, Bansal, Han, and Chang}]{yin2023dynosaur}
Da~Yin, Xiao Liu, Fan Yin, Ming Zhong, Hritik Bansal, Jiawei Han, and Kai-Wei Chang. 2023.
\newblock Dynosaur: A dynamic growth paradigm for instruction-tuning data curation.
\newblock \emph{arXiv preprint arXiv:2305.14327}.

\bibitem[{Zhou et~al.(2024)Zhou, Liu, Xu, Iyer, Sun, Mao, Ma, Efrat, Yu, Yu et~al.}]{zhou2024lima}
Chunting Zhou, Pengfei Liu, Puxin Xu, Srinivasan Iyer, Jiao Sun, Yuning Mao, Xuezhe Ma, Avia Efrat, Ping Yu, Lili Yu, et~al. 2024.
\newblock Lima: Less is more for alignment.
\newblock \emph{Advances in Neural Information Processing Systems}, 36.

\bibitem[{Zhou et~al.(2023)Zhou, Jiang, Wilcox, Cotterell, and Sachan}]{zhou2023controlled}
Wangchunshu Zhou, Yuchen~Eleanor Jiang, Ethan Wilcox, Ryan Cotterell, and Mrinmaya Sachan. 2023.
\newblock \href {https://arxiv.org/abs/2304.14293} {Controlled text generation with natural language instructions}.
\newblock \emph{Preprint}, arXiv:2304.14293.

\bibitem[{Ziegler et~al.(2020)Ziegler, Stiennon, Wu, Brown, Radford, Amodei, Christiano, and Irving}]{ziegler2020finetuning}
Daniel~M. Ziegler, Nisan Stiennon, Jeffrey Wu, Tom~B. Brown, Alec Radford, Dario Amodei, Paul Christiano, and Geoffrey Irving. 2020.
\newblock \href {https://arxiv.org/abs/1909.08593} {Fine-tuning language models from human preferences}.
\newblock \emph{Preprint}, arXiv:1909.08593.

\end{thebibliography}
